\documentclass{article}

\usepackage{microtype}
\usepackage{graphicx}
\usepackage{enumitem}
\usepackage{subfigure}
\usepackage{booktabs} 

\usepackage{hyperref}

\usepackage[accepted]{icml2019}
\usepackage{pifont}\newcommand{\cmark}{\ding{51}}\newcommand{\xmark}{\ding{55}}

\usepackage{amsmath,amsfonts,bm}

\newcommand{\mb}{\mathbf}

\def\eqref#1{equation~\ref{#1}}

\def\1{\bm{1}}

\DeclareMathAlphabet{\mathsfit}{\encodingdefault}{\sfdefault}{m}{sl}
\SetMathAlphabet{\mathsfit}{bold}{\encodingdefault}{\sfdefault}{bx}{n}

\usepackage{hyperref}
\usepackage{url}
\newcommand{\xhdr}[1]{{\noindent\bfseries #1}.}
\usepackage{algorithm}
\usepackage{xcolor}

\newcommand{\cut}[1]{}

\newcommand{\name}{GraphLog}

\icmltitlerunning{Evaluating Logical Generalization in Graph Neural Networks}
\begin{document}

\twocolumn[
\icmltitle{Evaluating Logical Generalization in Graph Neural Networks}

\icmlsetsymbol{equal}{*}

\begin{icmlauthorlist}
\icmlauthor{Koustuv Sinha}{fb,mcgill,mila}
\icmlauthor{Shagun Sodhani}{fb}
\icmlauthor{Joelle Pineau}{fb,mcgill,mila}
\icmlauthor{William L. Hamilton}{mcgill,mila}
\end{icmlauthorlist}

\icmlaffiliation{fb}{Facebook AI Research, Montreal, Canada}
\icmlaffiliation{mcgill}{School of Computer Science, McGill University, Montreal, Canada}
\icmlaffiliation{mila}{Montreal Institute of Learning Algorithms (Mila)}

\icmlcorrespondingauthor{Koustuv Sinha}{koustuv.sinha@mail.mcgill.ca}

\icmlkeywords{Machine Learning, ICML}

\vskip 0.3in
]

\printAffiliationsAndNotice{\icmlEqualContribution} 

\begin{abstract}
Recent research has highlighted the role of relational inductive biases in building learning agents that can generalize and reason in a compositional manner. 
However, while relational learning algorithms such as graph neural networks (GNNs) show promise, we do not understand how effectively these approaches can adapt to new tasks. In this work, we study the task of {\em logical generalization} using GNNs by designing a benchmark suite grounded in first-order logic. 
Our benchmark suite, \name, requires that learning algorithms perform rule induction in different synthetic logics, represented as knowledge graphs. 
\name\ consists of relation prediction tasks on 57 distinct logical domains.
We use \name\ to evaluate GNNs in three different setups: single-task supervised learning, multi-task pretraining, and continual learning. 
Unlike previous benchmarks, our approach allows us to precisely control the logical relationship between the different tasks.
We find that the ability for models to generalize and adapt is strongly determined by the diversity of the logical rules they encounter during training, and our results highlight new challenges for the design of GNN models. We publicly release the dataset and code used to generate and interact with the dataset at \href{https://www.cs.mcgill.ca/~ksinha4/graphlog/}{https://www.cs.mcgill.ca/~ksinha4/graphlog/}.
\end{abstract}

\section{Introduction}

Relational reasoning, or the ability to reason about the relationship between objects entities in the environment, is considered a fundamental aspect of intelligence \citep{2011-a-hierarchy-for-relational-reasoning-in-the-prefrontal-cortex, 2010-relational-knowledge-the-foundation-of-higher-cognition}. Relational reasoning is known to play a critical role in cognitive growth of children \citep{2011-connecting-instances-to-promote-childrens-relational-reasoning, 2007-the-role-of-relational-reasoning-in-childrens-addition-concept, 2010-young-childrens-analogical-reasoning-across-cultures-similarities-and-differences}. This ability to infer relations between objects/entities/situations, and to compose relations into higher-order relations, is one of the reasons why humans quickly learn how to solve new tasks \citep{2012-the-oxford-handbook-of-thinking-and-reasoning, 2016-relational-thinking-and-relational-reasoning-harnessing-the-power-of-patterning}.

\begin{figure}
    \centering
    \includegraphics[trim=100 150 220 50,clip,width=0.35\textwidth]{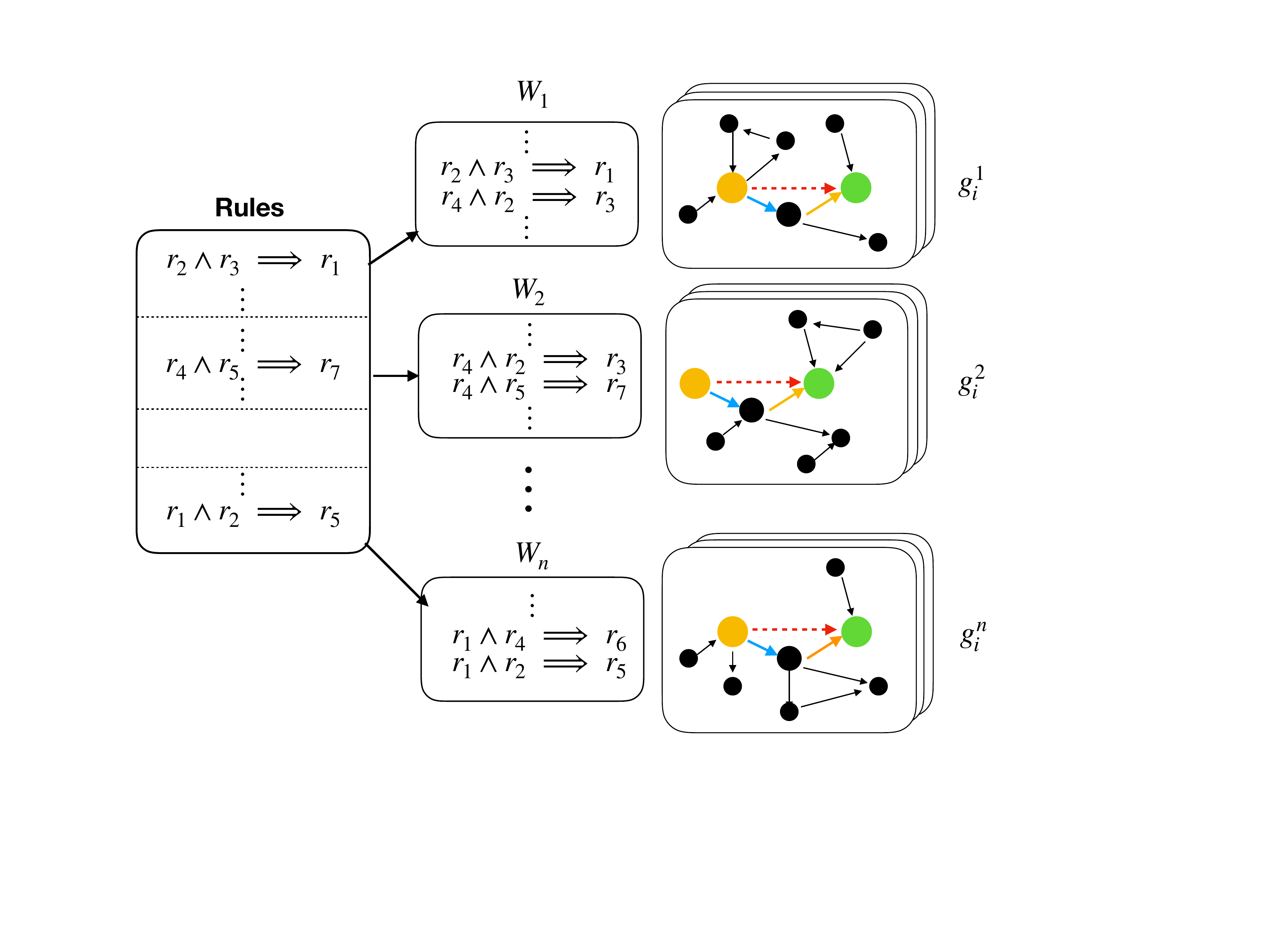}
    \caption{
    \name{} setup. We define a large set of rules that are grounded in propositional logic. We partition the rule set into overlapping subsets, which we use to define the unique {\em worlds}, $W_k$. Finally, within each \textit{world} $W_k$, we generate several knowledge graphs $g_i^k$ that are governed by the rule set of $W_k$.}
    \label{fig:worlds}
\end{figure}

The perceived importance of relational reasoning for generalization capabilities has fueled the development of several neural network architectures that incorporate relational inductive biases \citep{2016-interaction-networks-for-learning-about-objects-relations-and-physics, 2017-a-simple-neural-network-module-for-relational-reasoning, 2018-relational-inductive-biases-deep-learning-and-graph-networks}.
Graph neural networks (GNNs), in particular, have emerged as a dominant computational paradigm within this growing area \cite{2008_the_gnn_model,hamilton2017representation, 2017_neural_message_passing, 2018_rgcn, 2019_graph_neural_tangent_kernel_fusing_graph_neural_networks_with_graph_kernels}. 
However, despite the growing interest in GNNs and their promise for improving the generalization capabilities of neural networks, we currently lack an understanding of how \textit{effectively} these models can adapt and generalize across distinct tasks. 

In this work, we study the task of {\em logical generalization}, in the context of relational reasoning using GNNs. 
In particular, we study how GNNs can induce logical rules and generalize by combining these rules in novel ways after training. 
We propose a benchmark suite, \name, that is grounded in first-order logic. Figure \ref{fig:worlds} shows the setup of the benchmark. Given a set of logical rules, we create different logical worlds with overlapping rules. For each world (say $W_k$), we sample multiple knowledge graphs (say $g_i^k$). The learning agent should learn to induce the logical rules for predicting the missing facts in these knowledge graphs.
Using our benchmark, we evaluate the generalization capabilities of GNNs in a supervised setting by predicting unseen combinations of known rules within a specific logical \textit{world}. This task that explicitly requires inductive generalization.
We further analyze how various GNN architectures perform in the multi-task and the continual learning scenarios, where they have to learn over a set of logical worlds with different underlying logic.
Our setup allows us to control the similarity between the different \textit{worlds} by controlling the overlap in logical rules between different \textit{worlds}. This enables us to precisely analyze how task similarity impacts performance in the multi-task setting. 

Our analysis provides the following useful insights regarding the logical generalization capabilities of GNNs:
\begin{itemize}[itemsep=3pt,topsep=0pt,parsep=0pt]
\item Two architecture choices for GNNs have a strong positive impact on the generalization performance: \textbf{1)} incorporating multi-relational edge features using attention, and \textbf{2)} explicitly modularising the GNN architecture to include a parametric {\em representation function}, which learns representations for the relations based on the knowledge graph structure. 
    \item In the multi-task setting, training a model on a more diverse set of logical \textit{worlds} improves generalization and adaptation performance.
    
    \item All the evaluated models exhibit catastrophic forgetting in the continual learning setting. This indicates that the models are prone to fitting to just the current task at hand and not learning representations and compositions that can transfer across tasks---highlighting the challenge of lifelong learning in the context of logical generalization and GNNs.
\end{itemize}

\section{Background and Related Work}
\label{sec:background}
\xhdr{Graph Neural Networks}
Several graph neural network (\textit{GNN}) architectures have been proposed to learn the representation for the graph input \cite{2008_the_gnn_model, 2015_convolutional_networks_on_graphs_for_learning_molecular_fingerprints, 2016_convolutional_networks_on_graphs_with_fast_localized_spectral_filtering, 2016_gcn, 2017_neural_message_passing, 2017_graph_attention_network, 2017_inductive_representation_learning_on_large_graphs, 2018_rgcn}. Previous works have focused on evaluating graph neural networks in terms of their expressive power \cite{morris2019weisfeiler,2018_how_powerful_are_graph_neural_networks}, usefulness of features \cite{2019_are_powerful_graph_neural_nets_necessary_a_dissection_on_graph_classification}, and explaining the predictions from GNNs \cite{2019_gnnexplainer_generating_explanations_for_graph_neural_networks}. Complementing these works, we evaluate GNN models on the task of logical generalization.

\xhdr{Knowledge graph completion}
Many knowledge graph datasets are available for the task of relation prediction (also known as knowledge base completion). Prominent examples include Freebase15K \cite{2013_translating_embeddings_for_modeling_multi_relational_data}, WordNet \cite{1995_wordnet_a_lexical_database_for_english}, NELL \cite{never_ending_language_learning}, and YAGO \cite{2007_yago_a_core_of_semantic_knowledge, 2011_yago2_exploring_and_querying_world_knowledge_in_time_space_context_and_many_languages, 2013_yago3_a_knowledge_base_from_multilingual_wikipedias}. These datasets are derived from real-world knowledge graphs and are useful for empirical evaluation of relation prediction systems. However, these datasets are generally noisy and incomplete, as many facts are not available in the underlying knowledge bases \cite{west2014knowledge, paulheim2017knowledge}.
Moreover, the logical rules underpinning these systems are often opaque and implicit \cite{guo2016jointly}. All these shortcomings reduce the usefulness of existing knowledge graph datasets for understanding the logical generalization capability of neural networks. Some of these limitations can be overcome by using synthetic datasets, which can provide a high degree of control and flexibility over the data generation process at a low cost. Synthetic datasets are useful for understanding the behavior of different models - especially when the underlying problem can have many factors of variations. We consider using synthetic datasets, as a means and not an end, to understand the logical generalization capability of GNNs. 

Our \name\ benchmark serves as a synthetic complement to the real-world datasets. Instead of sampling from a real-world knowledge base, we create synthetic knowledge graphs that are governed by a known and inspectable set of logical rules. Moreover, the relations in \name\ are self-contained and do not require any common-sense knowledge, thus making the tasks self-contained.

\xhdr{Procedurally generated datasets for reasoning}
In recent years, several procedurally generated benchmarks have been proposed to study the relational reasoning and compositional generalization properties of neural networks. Some recent and prominent examples are listed in Table \ref{tab:dataset_comparison}. These datasets aim to provide a controlled testbed for evaluating the compositional reasoning capabilities of neural networks in isolation. Based on these existing works and their insightful observations, we enumerate the four key desiderata that, we believe, such a benchmark should provide:

\begin{enumerate}[itemsep=1pt,topsep=0pt, parsep=0pt, leftmargin=*]
    \item {\bf Interpretable Rules:} The rules that are used to procedurally generate the dataset should be human interpretable. 
    \item {\bf Diversity:} The benchmark datasets should have enough diversity across different tasks, and the compositional rules used to solve different tasks should be distinct, so that adaptation on a novel task is not trivial. The degree of similarity across the tasks should be configurable to enable evaluating the role of diversity in generalization.
    \item {\bf Compositional generalization: }The benchmark should require compositional generalization, i.e., generalization to unseen combinations of rules.
    \item {\bf Number of tasks: }The benchmark should support creating a large number of tasks. This enables a more fine-grained inspection of the generalization capabilities of the model in different setups, e.g., supervised learning, multitask learning, and continual learning.
\end{enumerate}
As shown in Table \ref{tab:dataset_comparison}, \name\ is unique in satisfying all of these desiderata. We highlight that \textbf{\name\ is the only dataset specifically designed to test logical generalization capabilities on graph data}, whereas previous works have largely focused on the image and text modalities. 

\begin{table}[]
\centering
\resizebox{0.5\textwidth}{!}{%
\begin{tabular}{|lllllllll|}
\hline
Dataset & \textbf{IR} & \textbf{D} & \textbf{CG} & \textbf{M} & \textbf{S} & \textbf{Me} & \textbf{Mu} & \textbf{CL} \\ \hline
CLEVR \cite{johnson2017clevr} & \cmark & \xmark & \xmark& Vision & \cmark& \xmark & \xmark & \xmark \\
CoGenT \cite{johnson2017clevr} & \cmark& \xmark& \cmark& Vision & \cmark& \xmark & \xmark & \xmark \\
CLUTRR \cite{sinha2019clutrr} & \cmark& \xmark& \cmark& Text & \cmark& \xmark & \xmark & \xmark \\
SCAN \cite{lake2017generalization} & \cmark& \xmark&  \cmark& Text & \cmark&  \cmark&  \xmark & \xmark \\
SQoOP \cite{bahdanau2018systematic} & \cmark& \xmark& \cmark& Vision & \cmark& \xmark & \xmark & \xmark \\
TextWorld \cite{cote2018textworld} & \xmark& \cmark& \cmark& Text & \cmark& \cmark& \cmark& \cmark \\
GraphLog (Proposed) & \cmark& \cmark&  \cmark& Graph & \cmark& \cmark& \cmark& \cmark \\ \hline
\end{tabular}%
}
\caption{Features of related datasets that are: 1) designed to test compositional generalization and reasoning, and 2) procedurally gnerated. We compare the datasets along the following dimensions: Inspectable Rules (\textbf{IR}), \textbf{D}iversity, Compositional Generalization (\textbf{CG}), \textbf{M}odality and if the following training setups are supported: \textbf{S}upervised, \textbf{Me}ta-learning, \textbf{Mu}ltitask \& Continual learning (\textbf{CL}).}
\label{tab:dataset_comparison}
\end{table}

\section{GraphLog}
\label{sec:graph_log}

\subsection{Terminology}
\label{sec:terminology}

A \textit{graph} $G = (V_G, E_G)$ is a collection of a set of nodes $V_G$ and a set of edges $E_G$ between the nodes.
In this work, we assume that each pair of nodes have at most one edge between them. A \textit{relational graph} is a graph where the edge between two nodes (say $u$ and $v$) is assigned a \textit{label}, denoted \textit{$r$}. The labeled edge is denoted as $(u \rightarrow_r v) \in E_G$.
A \textit{relation set} $R$ is a set of relations $\{r_1$, $r_2$, ... $r_K\}$. A \textit{rule set} $\mathcal{R}$ is a set of rules in first order logic, which we denote in the Datalog format \cite{evansLearningExplanatoryRules2017}, $[r_i, r_j] \Rightarrow r_k$, and which can be expanded as {\em Horn clauses} of the form:
\begin{equation}\label{eq:rule}
    \exists z \in V_G : (u \rightarrow_{r_i} z) \land  (z \rightarrow_{r_j} v) \Rightarrow   (u \rightarrow_{r_k} v) 
\end{equation}
where $z$ denotes a variable that can be bound to any entity and $\Rightarrow$ denotes logical implication. 
The relations $r_i, r_j$ form the \textit{body} while the relation $r_k$ forms the \textit{head} of the rule. 
Horn clauses of this form represent a well-defined subset of first-order logic, and they encompass the types of logical rules learned by the vast majority of existing rule induction engines for knowledge graphs \cite{langley1995applications}.

We use 
$p_G^{u, v}$  to denote a \textit{path} from node $u$ to $v$ in a graph $G$. 
We construct graphs according to rules of the form in Equation \ref{eq:rule} so that a path between two nodes will always imply a specific relation between these two nodes.
In other words, we will always have that
\begin{equation}
    \exists r_i \in \mathcal{R} \::\: p_G^{u, v} \Rightarrow (u \rightarrow_{r_i} v).
\end{equation}
Thus, following the path between two nodes, and applying the propositional rules along the edges of the path, we can \textit{resolve} the relationship between the nodes. Hence, we refer to the paths as \textit{resolution paths}. The edges of the resolution path are concatenated together to obtain a \textit{descriptor}. These descriptors are used for quantifying the similarity between different resolution paths, with a higher overlap between the descriptors implying a greater similarity between two resolution paths.

\subsection{Problem Setup}
\label{sec:problem_setup}

We formulate the relational reasoning task as predicting relations between the nodes in a relational graph. Given a query  $(G, u, v)$ where $u, v \in V_G$, the learner has to predict the relation $r_?$ for the edge $u \rightarrow_{r_?} v$. Unlike the previous work on knowledge graph completion, we emphasize an {\em inductive} problem setup, where the graph $G$ in each query is unique. Rather than reasoning on a single static knowledge graph during training and testing, we consider the setting where the model must learn to generalize to unseen graphs during evaluation.

\subsection{Dataset Generation}
\label{sec:dataset_generation}

As discussed in Section \ref{sec:background}, we want our proposed benchmark to provide four key desiderata: (i) interpretable rules, (ii) diversity, (iii) compositional generalization and (iv) large number of tasks. We describe how our dataset generation process ensures all four aspects.

\xhdr{Rule generation} We create a set $R$ of $K$ relations and use it to sample a rule set $\mathcal{R}$. We impose two constraints on $\mathcal{R}$: \textbf{(i)} No two rules in $\mathcal{R}$ can have the same body. This ensures consistency between the rules. \textbf{(ii)} Rules cannot have common relations among the \textit{head} and \textit{body}. This ensures the absence of cyclic dependencies in rules \cite{NIPS2018_7473}. Generating the dataset using a consistent and well-defined rule set ensures interpretability in the resulting dataset. The full algorithm for rule generation is given in Appendix (Algorithm \ref{alg:rule_generation}).

\xhdr{Graph generation} The graph generation process has two steps: In the first step, we recursively sample and use rules in $\mathcal{R}$ to generate a relational graph called the \textit{WorldGraph} (as shown in Figure \ref{fig:worlds}). This sampling procedure enables us to create a diverse set of \textit{WorldGraphs} by considering only certain subsets (of $\mathcal{R}$) during sampling. By controlling the extent of overlap between the subsets of $\mathcal{R}$ (in terms of the number of rules that are common across the subsets), we can precisely control the similarity between the different \textit{WorldGraphs}. The full algorithm for generating the \textit{WorldGraph} and controlling the similarity between the \textit{worlds} is given in Appendix (Algorithm \ref{alg:world_graph} and Section \ref{sec:world_sampling_app}).

In the second step, the \textit{WorldGraph} $G_W$ is used to sample a set of graphs $G_W^S = (g_1, \cdots g_N)$ (shown as Step (a) in Figure \ref{fig:overview}). A graph $g_i$ is sampled from $G_W$ by sampling a pair of nodes $(u, v)$ from $G_W$ and then by sampling a resolution path $p_{G_W}^{u, v}$. The edge $u \rightarrow_{r_i} v$ between the source and sink node of the path provides the target relation for the learning model to predict. 
To increase the complexity of the sampled $g_i$ graphs (beyond being simple paths), we also add nodes to $g_i$ by sampling neighbors of the nodes on $p_{G_W}^{u, v}$, such that no other shortest path exists between $u$ and $v$. Algorithm \ref{alg:sample_graph} (in the Appendix) details our graph sampling approach.

\subsection{Summary of the \name\ Dataset}

We use the data generation process described in Section \ref{sec:dataset_generation} to instantiate a dataset suite with 57 distinct logical \textit{worlds} and $5000$ graphs per \textit{world} (Figure \ref{fig:worlds}). 
The dataset is divided into the sets of training, validation, and testing \textit{worlds}. The graphs within each \textit{world} are also split into training, validation, and testing sets. The key statistics of the datasets are given in Table \ref{tab:data_agg_stats}. Though we instantiate 57 \textit{worlds}, the \name\ code can instantiate an arbitrary number of \textit{worlds} and has been included in the supplementary material.

\subsubsection{Setups supported in \name}

\name\ enables us to investigate the logical relational reasoning performance of models in the following setups:

\begin{table}[]
\centering
{\small
\begin{tabular}{|ll|}
\hline 
Number of relations & 20 \\
Total number of \textit{WorldGraph}s & 57 \\
Total number of unique rules & 76 \\
Training Graphs per \textit{WorldGraph} & 5000 \\
Validation Graphs per \textit{WorldGraph} & 1000 \\
Testing Graphs per \textit{WorldGraph} & 1000 \\
Number of rules per \textit{WorldGraph} & 20 \\
Average number of \textit{descriptors} & 522 \\
Maximum length of resolution path & 10 \\
Minimum length of resolution path & 2 \\ \hline
\end{tabular}%
}
\caption{Aggregate statistics of the worlds used in \name{}. Statistics for each individual world are in the Appendix. 
}
\label{tab:data_agg_stats}
\end{table}

\xhdr{Supervised learning} 
In the supervised learning setup, a model is trained on the train split of a single logical \textit{world} and evaluated on the test split of the same world. The total number of rules grows exponentially with the number of relations $K$, making it impossible to train on all possible combinations of the relations. However, we expect that a \textit{perfectly} systematic model generalizes to unseen combinations of relations by training only on a subset of combinations (i.e., via inductive reasoning).

\xhdr{Multi-task learning}
\name\ provides multiple logical \textit{worlds}, each with its own training and evaluation splits. In the standard multi-task training, the model is trained on the train split of many \textit{worlds} ($W_1, \cdots, W_M)$ and evaluated on the test split of the same worlds. The complexity of each world and the similarity between the different worlds can be precisely controlled. \name\ thus enables us to evaluate how model performance varies when the model is trained on similar vs.\@ dissimilar \textit{worlds}.

\name\ is also designed to study the effect of pre-training on adaptation. In this setup, the model is first pre-trained on the train split of multiple \textit{worlds} ($W_1, \cdots, W_M$) and then adapted (fine-tuned) on the train split of the unseen heldout \textit{worlds} ($W_{M+1}, \cdots, W_N)$. The model is evaluated on the \textit{test} split of the novel \textit{worlds}. Similar to the previous setup, \name\ provides us an opportunity to investigate the effect of \textit{similarity} in pre-training. This enables \name\ to mimic in-distribution and out-of-distribution training and testing scenarios, as well as precisely categorize the effect of multi-task pre-training for adaptation performance.

\xhdr{Continual learning}
\name\ provides access to a large number of \textit{worlds}, enabling us to evaluate the logical generalization capability of the models in the continual learning setup. In this setup, the model is trained on a sequence of \textit{worlds}. Before training on a new \textit{world}, the model is evaluated on all the \textit{worlds} that the model has trained on so far.
Given the several challenges involved in continual learning \cite{2012_learning_to_learn, 2019_continual_lifelong_learning_with_neural_networks_a_review, 2019_continual_learning_a_comparative_study_on_how_to_defy_forgetting_in_classification_tasks, 2019_on_training_recurrent_neural_networks_for_lifelong_learning}, we do not expect the models to be able to remember the knowledge from all the previous tasks. Nonetheless, given that we are evaluating the models for relational reasoning and that our datasets share \textit{relations}, we would expect the models to retain some knowledge of how to solve the previous tasks. 
In this sense, the performance on the previous tasks can also be seen as an indicator of if the models actually learn to solve the relational reasoning tasks or they just \textit{fit} to the current dataset distribution.

\begin{figure*}[h]
    \centering
    \includegraphics[trim=10 190 50 250,clip,width=0.95\textwidth]{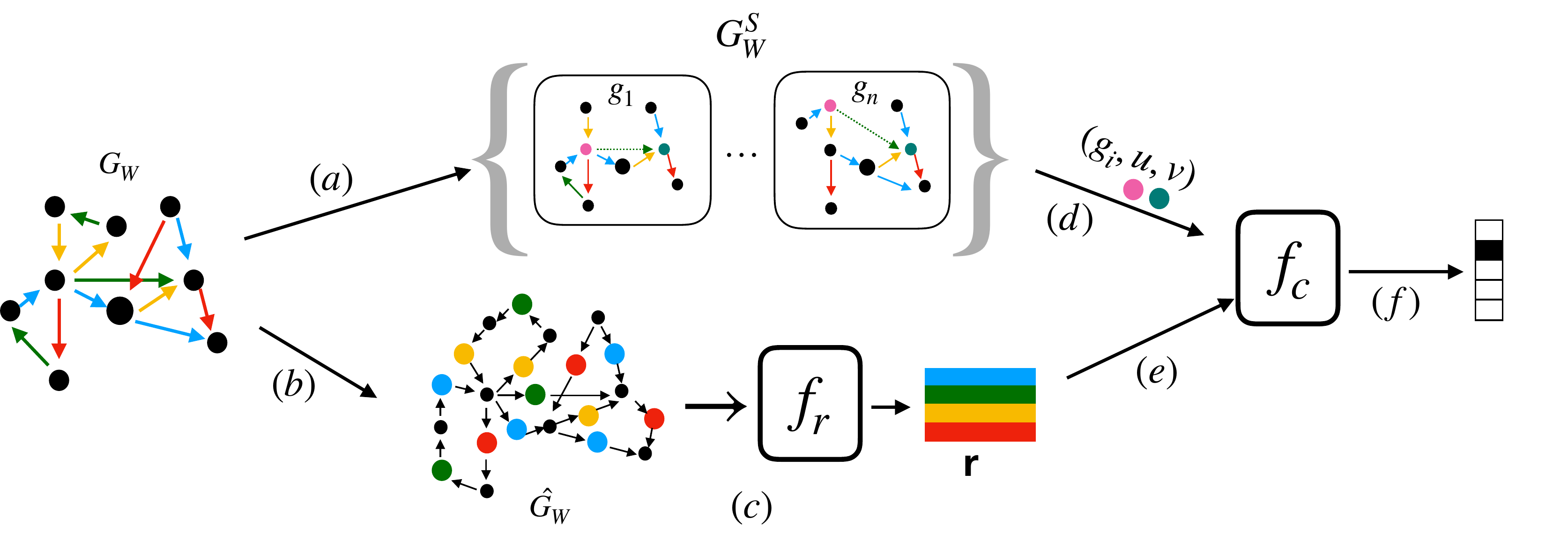}
    \vspace{-7pt}
\caption{Overview of the training process: \textbf{(a)}: Sampling multiple graphs from $G_W$. \textbf{(b)}: Converting the relational graph into extended graph $\hat{G_W}$. Note that edges of different color (denoting different types of relations) are replaced by a node of same type in $\hat{G}_{W}$. \textbf{(c)}: Learning representations of the relations ($\mb{r}$) using $f_r$ with the extended graph as the input. In case of \texttt{Param} models, the relation representations are parameterized via an embedding layer and the extended graph is not created. \textbf{(d, e)}: The composition function takes as input the query $g_i, u, v$ and the relational representation $\mb{r}$. \textbf{(f)}: The composition function predicts the relation between the nodes $u$ and $v$.}
\label{fig:overview}
\end{figure*}

\section{Representation and Composition}\label{sec:comp_rep}
In this section, we describe the graph neural network (GNN) architectures that we evaluate on the \name\ benchmark. 
In order to perform well on the benchmark tasks, a model should learn representations that are useful for solving the tasks in the current \textit{world} while being general enough to be effectively adapted to the new \textit{worlds}. 
To this end, we structure the GNN models we analyze around two key modules:
\begin{itemize}[itemsep=0pt,parsep=0pt,topsep=0pt,leftmargin=*]
    \item \textbf{Representation module:} This module is represented as a function $f_r : G_{W} \times R \rightarrow \mathbb{R}^d$, which maps logical relations within a particular \textit{world} $W$ to $d$-dimensional vector representations. 
    Intuitively, this function should learn how to encode the {\em semantics} of the various relations within a logical \textit{world}.
    \item \textbf{Composition module: } This module is a function $f_c : G_{W} \times V_{G_{W}} \times V_{G_{W}} \times \mathbb{R}^{d \times |R|} \rightarrow R$, which learns how to compose the relation representations learned by $f_r$ to make predictions about queries over a knowledge graph. 
\end{itemize}
Note that though we break down the process into two steps, in practice, the learner does not have access to the \textit{correct} representations of relations or to $\mathcal{R}$. The learner has to rely only on the target labels to solve the reasoning task. We hypothesize that this \textit{separation of concerns} between a \textit{representation function} and a \textit{composition function} \cite{1982_on_the_role_of_scientific_thought} could provide a useful inductive bias for the model.

\subsection{Representation modules}
We first describe the different approaches for learning the representation  $\mb{r}_i \in \mathbb{R}^d$ for the relations. These representations will be provided as input to the \textit{composition function}.

\xhdr{Direct parameterization}
The simplest approach to define the representation module is to train unique embeddings for each relation $r_i$. This approach is predominantly used in the previous work on GNNs \cite{2017_neural_message_passing, 2017_graph_attention_network}, and we term this approach as the {\em Param} representation module. 
A major limitation of this approach is that the relation representations are optimized specifically for each logical world, and there is no inductive bias towards learning representations that can generalize.

\xhdr{Learning representations from the graph structure}
In order to define a more powerful and expressive representation function, we consider an approach that learns relation representations as a function of the \textit{WorldGraph} underlying a logical world.
To do so, we consider an ``extended'' form of the \textit{WorldGraph}, $\hat{G}_W$, where introduce new nodes (called \textit{edge-nodes}) corresponding to each edge in the original \textit{WorldGraph} $G_W$. For an edge $(u \rightarrow_r v) \in E_G$, the corresponding edge-node $(u-r-v)$ is connected to only those nodes that were incident to it in the original graph (i.e. nodes $u$ and $v$; see Figure \ref{fig:overview}, Step (b)). This new graph $\hat{G}_W$ only has one type of edge and comprises of nodes from both the original graph and from the set of edge-nodes.

We learn the relation representations by training a GNN model on the expanded \textit{WorldGraph} and by averaging the edge-node embeddings corresponding to each relation type $r_i \in R$. (Step (c) in Figure \ref{fig:overview}).
For the GNN model, we consider the Graph Convolutional Network (\texttt{GCN}) \cite{2016_gcn} and the Graph Attention Network (\texttt{GAT}) architectures. 
Since the nodes do not have any features or attributes, we randomly initialize the embeddings in these GNN message passing layers. 

The intuition behind creating the extended-graph is that the representation GNN function can learn the relation embeddings based on the structure of the complete relational graph $G_W$. We expect this to provide an inductive bias that can generalize more effectively than the simple \textit{Param} approach.
Finally, note that while the representation function is given access to the \textit{WorldGraph} to learn representations for relations, the composition module is not able to interface with the \textit{WorldGraph} in order to make predictions about a query.

\subsection{Composition modules}

We now describe the GNNs used for the composition modules. These models take as input the query $(g_i, u, v)$ and the relation embedding $\mb{r}_i \in \mathbb{R}^d$ (Step (d) and (e) in Figure \ref{fig:overview}).

\xhdr{Relational Graph Convolutional Network (RGCN)}
Given that the input to the composition module is a relational graph, the RGCN model \cite{2018_rgcn} is a natural choice for a baseline architecture.
In this approach, we iterate a series of message passing operations:
\begin{equation*}
    \mb{h}_u^{(t)} = \textrm{ReLU}\left(\sum_{r_i \in R}\sum_{v \in \mathcal{N}_{r_i}(u)}\mb{r}_i\times_1\mathcal{T}\times_3\mb{h}_v^{(t-1)}\right),
\end{equation*}
where $\mb{h}^{(t)}_u \in \mathbb{R}^d$ denotes the representation for a node $u$ at the $t^{th}$ layer of the model, $\mathcal{T} \in \mathbb{R}^{d_r \times d \times d}$ is a learnable tensor, $\mb{r} \in \mathbb{R}^d$ is the representation for relation $r$, and $\mathcal{N}_{r_i}(u)$ denotes the neighbors of node $u$ by relation $r_i$.
We use $\times_i$ to denote multiplication across a particular mode of the tensor. 
This RGCN model learns a relation-specific propagation matrix, specified by the interaction between the relation embedding $\mb{r}_i$ and the shared tensor $\mathcal{T}$.\footnote{Note that the shared tensor is equivalent to the basis matrix formulation in \citet{2018_rgcn}.}

\xhdr{Edge-based Graph Attention Network (Edge-GAT)}
In addition to the RGCN model---which is considered the defacto standard architecture for applying GNNs to multi-relational data---we also explore an extension of the Graph Attention Network (GAT) model \cite{2017_graph_attention_network} to handle edge types. 
Many recent works have highlighted the importance of the attention mechanism, especially in the context of relational reasoning \cite{2017_attention_is_all_you_need, 2018_relational_recurrent_neural_networks, 2019_enhancing_the_transformer_with_explicit_relational_encoding_for_math_problem_solving}.
Motivated by this observation, we investigate an extended version of the GAT, where we incorporate gating via an LSTM \cite{hochreiter1997long} and where the attention is conditioned on both the incoming message (from the other nodes) and the relation embedding (of the other nodes):
\begin{align*}
    \mb{m}_{\mathcal{N}(u)} &= \sum_{r_i \in R}\sum_{v \in \mathcal{N}_{r_i}(u)}\alpha\left(\mb{h}_u^{(t-1)},\mb{h}^{(t-1)}_v, \mb{r}\right)\\
    \mb{h}_u^{(t)} &= \textrm{LSTM}(\mb{m}_{\mathcal{N}(u)}, \mb{h}_u^{(t-1)})
\end{align*}
Following the original \texttt{GAT} model, the attention function $\alpha$ is defined using an dense neural network on the concatenation of the input vectors.
We refer to this model as the Edge GAT (\texttt{E-GAT}) model. 

\xhdr{Query and node representations}
We predict the relation for a given query $(g_i,u,v)$ by concatenating $\mb{h}^{(K)}_u, \mb{h}^{(K)}_v$ (the final-layer query node embeddings, assuming a $K$-layer GNN) and applying a two-layer dense neural network (Step (f) in Figure \ref{fig:overview}). The entire model (i.e., the representation function and the composition function) are trained end-to-end using the softmax cross-entropy loss. 
Since we have no node features, we randomly initialize all the node embeddings in the GNNs (i.e., $\mb{h}_u^{(0)}$).

\section{Experiments}

\begin{figure}
    \centering
    \includegraphics[width=0.45\textwidth]{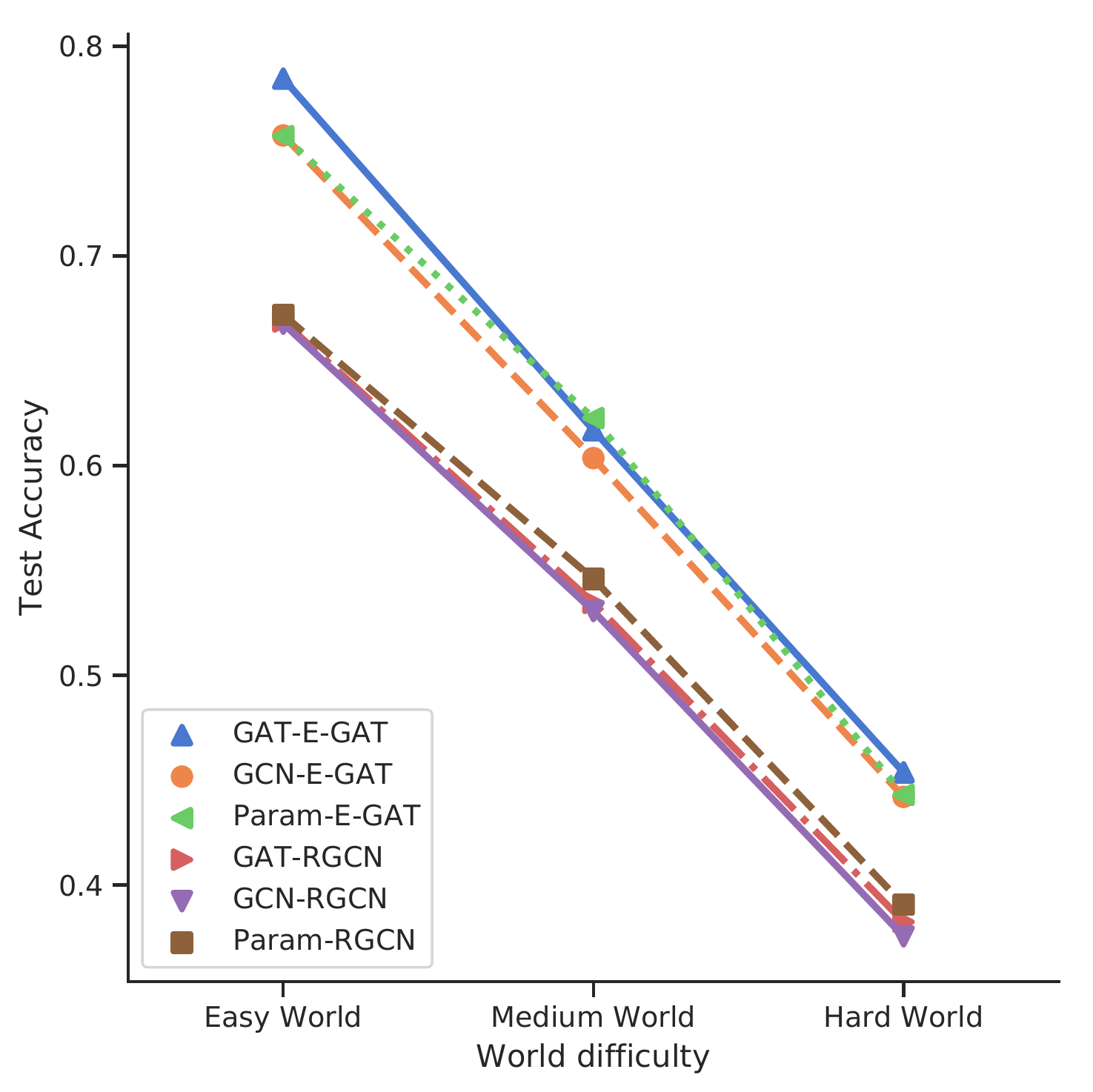}
    \vspace{-10pt}
    \caption{We categorize the datasets in terms of their relative \textit{difficulty} (see Appendix). We observe that the models using \texttt{E-GAT} as the composition function consistently work well.}
    \label{fig:inductive-reasoning}
\end{figure}

We aim to quantify the performance of the different GNN models on the task of logical relation reasoning, in three contexts: \textbf{(i)} Single Task Supervised Learning, \textbf{(ii)} Multi-Task Training and \textbf{(iii)} Continual Learning.
Our experiments use the \name\ benchmark with distinct 57 \textit{worlds} or knowledge graph datasets (see Section \ref{sec:graph_log}) and 6 different different GNN models (see Section \ref{sec:comp_rep}). In the main paper, we share the key trends and observations that hold across the different combinations of the models and the datasets, along with some representative results. The full set of results is provided in the Appendix. All the models are implemented using PyTorch 1.3.1 \cite{2019_pytorch}. The code has been included with the supplemental material.

\subsection{Single Task Supervised Learning}
\label{sec:single_task_supervised_learning_setup}
In our first setup, we train and evaluate all of the models on all the 57 \textit{worlds}, one model, and one \textit{world} pair at a time. This experiment provides several important results. Previous works considered only a handful of datasets when evaluating the different models on the task of relational reasoning. As such, it is possible to design a model that can exploit the biases present in the few datasets that the model is being evaluated over. In our case, we consider over 50 datasets, with different characteristics (Table~\ref{tab:data_agg_stats}). It is difficult for one model to outperform the other models on all the datasets just by exploiting some dataset-specific bias, thereby making the conclusions more robust.

In Figure \ref{fig:inductive-reasoning}, we present the results for the different models. We categorize the \textit{worlds} in three categories of \textit{difficulty} -- \textit{easy}, \textit{moderate} and \textit{difficult} -- based on relative test performance of the models on each \textit{world}. Table \ref{tab:full_supervised} (in Appendix) contains the results for the different models on the individual worlds. We observe that the models using \texttt{E-GAT} as the composition functions always outperform their counterparts using the \texttt{RGCN} models. This confirms our hypothesis about the usefulness of combining relational reasoning and attention for improving the performance on relational reasoning tasks. An interesting observation is that the relative ordering among the \textit{worlds}, in terms of the test accuracy of the different models, is consistent irrespective of the model we use, highlighting the intrinsic difficulty of the different {\em worlds} in \name.

\subsection{Multi-Task Training}
\label{sec:multi_task_training_setup}
\begin{table}[]
\centering
\resizebox{0.4\textwidth}{!}{%
\begin{tabular}{|ll|ll|l|}
\hline
 &  & \multicolumn{1}{c|}{S} & \multicolumn{1}{c|}{D} \\ \cline{1-4} 
$f_r$ & $f_c$ & Accuracy & Accuracy \\ \hline
GAT & E-GAT & \textbf{0.534} $\pm{0.11}$ & \textbf{0.534} $\pm{0.09}$ \\
GAT & RGCN & 0.474 $\pm{0.11}$ & 0.502 $\pm{0.09}$ \\\hline
GCN & E-GAT & 0.522 $\pm{0.1}$ & 0.533 $\pm{0.09}$ \\
GCN & RGCN & 0.448 $\pm{0.09}$ & 0.476 $\pm{0.09}$ \\ \hline
Param & E-GAT & 0.507 $\pm{0.09}$ & 0.5 $\pm{0.09}$ \\
Param & RGCN & 0.416 $\pm{0.07}$ & 0.449 $\pm{0.07}$ \\ \hline
\end{tabular}}%
\caption{Multitask evaluation performance when trained on different data distributions. We categorize the training distribution on basis of their similarity of rules: Similar (S) containing similar worlds and a mix of similar and dissimilar worlds (D)}
\label{tab:multitask_inductive}
\end{table}

We now turn to the setting of multi-task learning where we train the same model on multiple {\em logical worlds}.

\begin{figure}[h]
    \centering
    \includegraphics[width=0.45\textwidth]{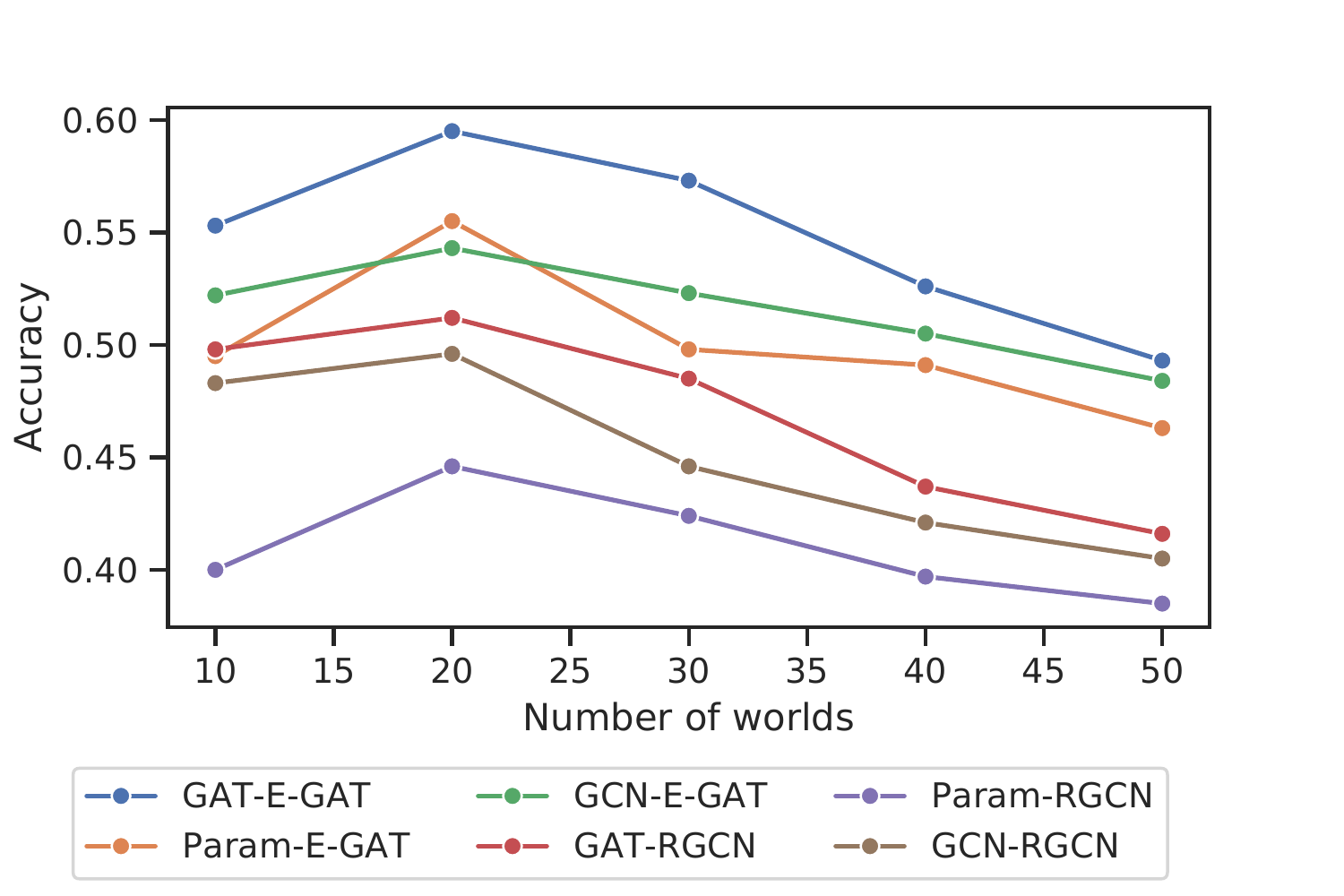}
    \vspace{-10pt}
    \caption{
    We run multitask experiments over an increasing number of worlds to stretch the capacity of the models. Evaluation performance is reported as the average of test set performance across the worlds that the model has trained on so far. All the models reach their optimal performance at 20 worlds, beyond which their performance starts to degrade.}
    \label{fig:mult_task_large}
\end{figure}

\xhdr{Basic multi-task training}
First, we evaluate a how changing the similarity among the training \textit{worlds} affects the test performance in the multi-task setup, where a model is trained jointly on eight and tested on three distinct {\em worlds.} In Table \ref{tab:multitask_inductive}, we observe that considering a mix of similar and dissimilar \textit{worlds} improves the generalization capabilities of all the models when evaluated on the test split. Another important observation is that just like the supervised learning setup, the \texttt{GAT-EGAT} model consistently performs either as good as or better than other models and the models using \texttt{EGAT} for the composition function perform better than the ones using the \texttt{RGCN} model.
Figure \ref{fig:mult_task_large} shows how the performance of the various models changes when we perform multi-task training on an increasingly large set of {\em worlds}.
Interestingly, we see that model performance improves as the number of {\em worlds} is increased from 10 to 20 but then begins to decline, indicating capacity saturation in the presence of too many diverse {\em worlds}.

\begin{figure}
    \centering
    \includegraphics[width=0.5\textwidth]{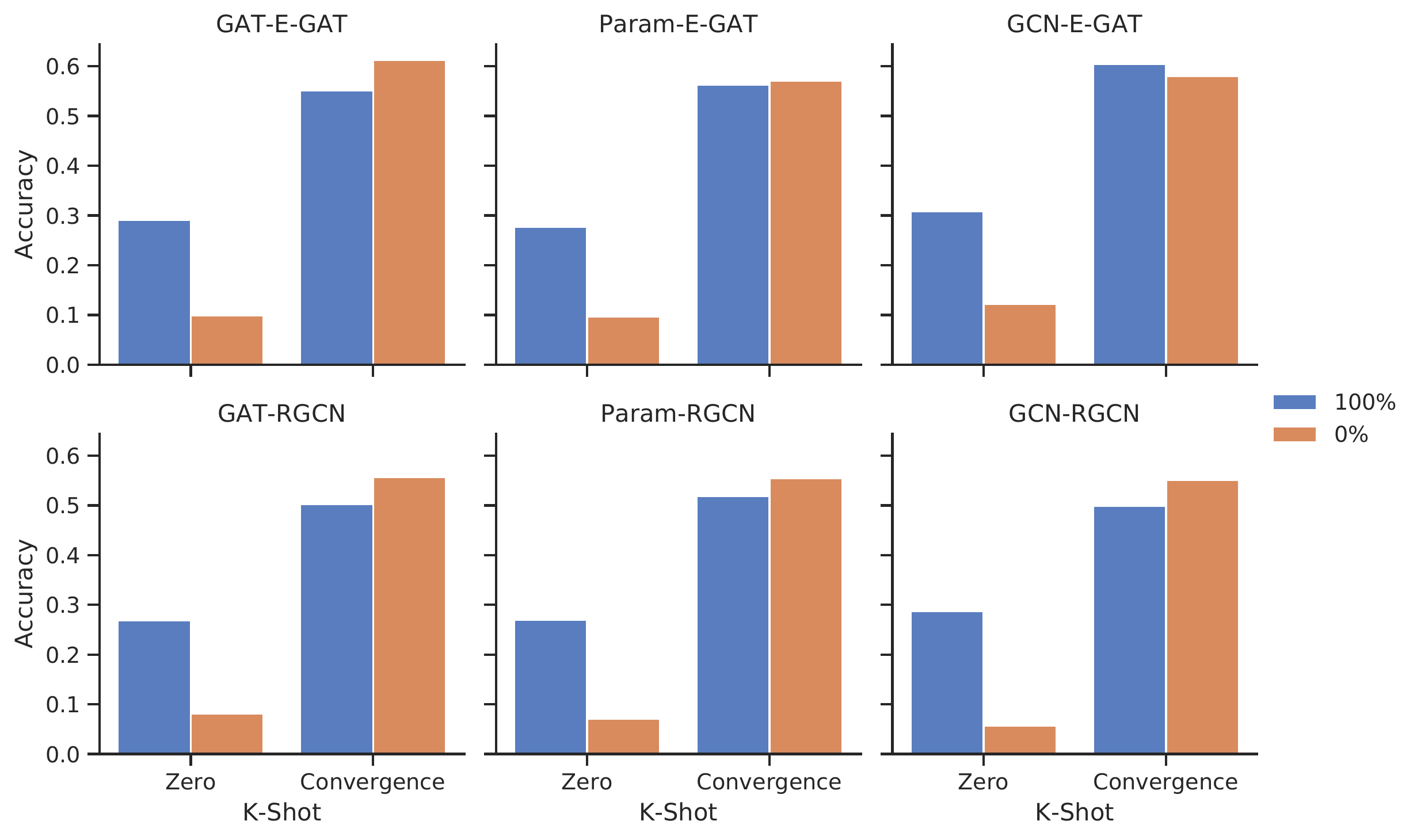}
    \vspace{-15pt}
    \caption{
    We evaluate the effect of changing the similarity between the training and the evaluation datasets. The colors of the bars depicts how similar the two distributions are while the \textit{y-axis} shows the mean accuracy of the model on the test split of the evaluation \textit{world}. We report both the zero-shot adaptation performance and performance after convergence.
}
    \label{fig:effect_of_similarity_between_train_and_eval}
\end{figure}

\xhdr{Multi-task pre-training}
In this setup, we pre-train the model on multiple \textit{worlds} and adapt on a heldout \textit{world}. We study how the models' adaption capabilities vary as the similarity between the training and the evaluation distributions changes. Figure \ref{fig:effect_of_similarity_between_train_and_eval} considers the case of zero-shot adaptation and adaptation till convergence. As we move along the \textit{x-axis}, the zero-shot performance (shown with solid colors) decreases in all the setups. This is expected as the similarity between the training and the evaluation distributions also decreases. An interesting trend is that the model's performance, after adaptation, increases as the similarity between the two distributions decreases. This suggests that training over a diverse set of distributions improves adaptation capability. The results for adaptation with 5, 10, ... 30 steps are provided in the Appendix (Figure \ref{fig:multitask_fine_grained}).

\begin{figure}[h]
    \centering
    \includegraphics[width=0.45\textwidth]{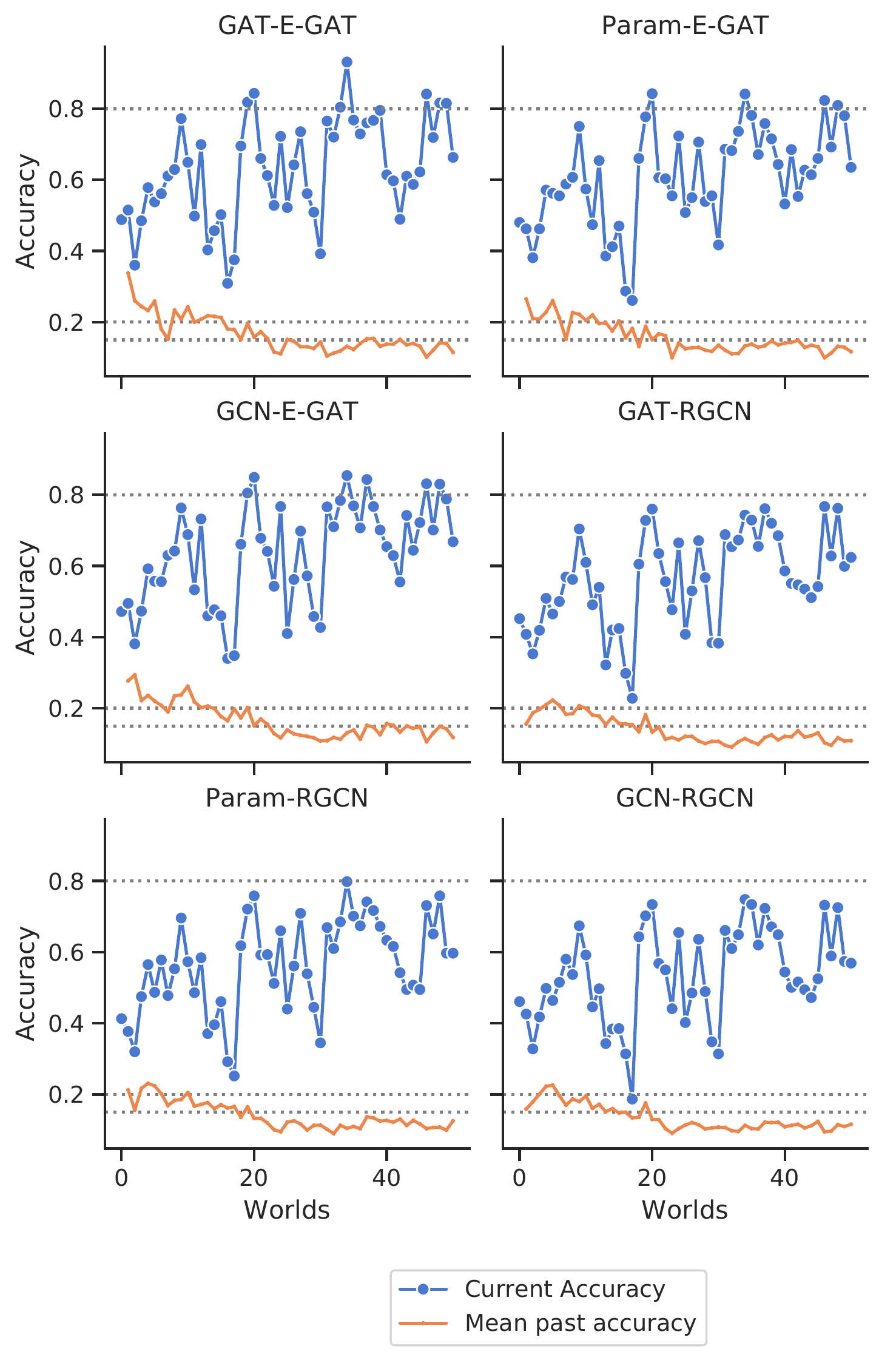}
    \vspace{-10pt}
    \caption{
    We evaluate the performance of all the models in a continual learning setup. The blue curve shows the accuracy on the current \textit{world} and the orange curve shows the mean accuracy on all the \textit{previously} seen \textit{worlds}. As the model trains on new \textit{worlds}, its performance on the previously seen \textit{worlds} degrades rapidly. This is the forgetting effect commonly encountered in continual learning setups.}
    \label{fig:catastrophic_forgetting}
\end{figure}

\begin{figure}[h]
    \centering
    \includegraphics[width=0.48\textwidth]{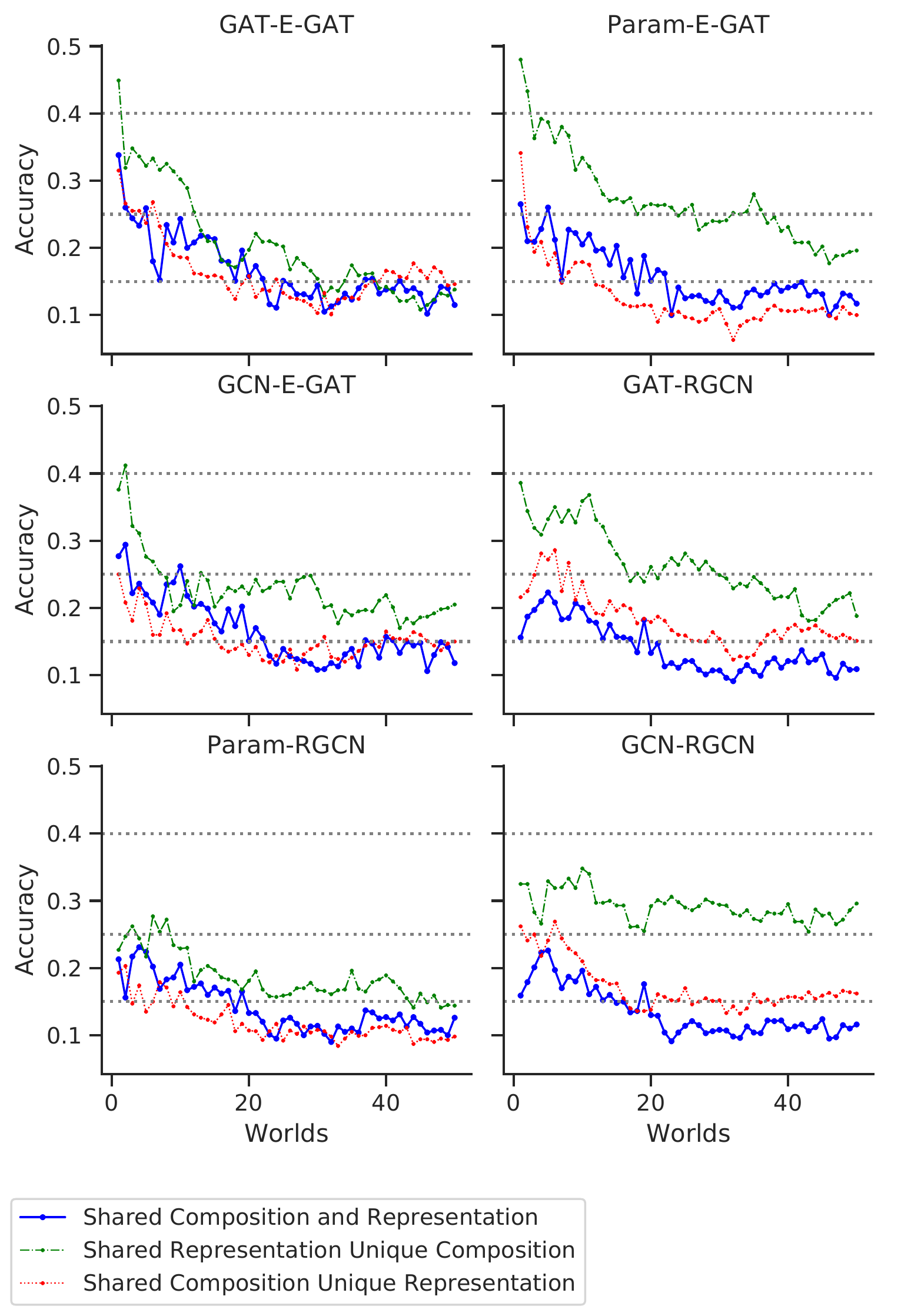}
    \vspace{-15pt}
    \caption{
    We evaluate the performance in a continual learning setup where we share either the representation function or the composition function or both. We observe that sharing the representation function reduces the effect of catastrophic forgetting as compared to the other setups.}
    \label{fig:catastrophic_forgetting_abalation}
\end{figure}

\subsection{Continual Learning Setup}
\label{sec:continual_learning}

In the continual learning setup, we evaluate the knowledge \textit{retention} capabilities of the GNN models. We train the model on a sequence of overlapping worlds, and after converging on every \textit{world}, we report the average of model's performance on all the previous \textit{worlds}.
In Figure \ref{fig:catastrophic_forgetting} we observe that as the model is trained on different \textit{worlds}, the performance on the previous \textit{worlds} degrades rapidly. This highlights that the current reasoning models are not suitable for continual learning.

\xhdr{The role of the representation function}
We also investigate 
the model's performance in a continual learning setup where the model learns only a \textit{world}-specific representation function or a \textit{world}-specific composition function, and where the other module is shared across the worlds. 
In Figure \ref{fig:catastrophic_forgetting_abalation}, we observe that sharing the representation function reduces the effect of catastrophic forgetting, but sharing the composition function does not have the same effect. This suggests that the representation function learns representations that are useful across the \textit{worlds}.

\section{Discussion \& Conclusion}

In this work, we propose \name, a benchmark suite for evaluating the logical generalization capabilities of Graph Neural Networks. \name\ is grounded in first-order logic and provides access to a large number of diverse tasks that require compositional generalization to solve, including single task supervised learning, multi-task learning, and continual learning.  
Our results highlight the importance of attention mechanisms and modularity to achieve logical generalization, while also highlighting open challenges related to multi-task and continual learning in the context of GNNs. 
A natural direction for future work is leveraging \name\ for studies of {\em fast adaptation} and {\em meta-learning} in the context of logical reasoning (e.g., via gradient-based meta learning), as well as integrating state-of-the-art methods (e.g., regularization techniques) to combat catastrophic forgetting in the context of GNNs.

\bibliography{paper}
\bibliographystyle{icml2019}

\appendix









\icmltitlerunning{Evaluating Logical Generalization in Graph Neural Networks}

\twocolumn[
\icmltitle{Supplemental Materials : Evaluating Logical Generalization in Graph Neural Networks}



\icmlsetsymbol{equal}{*}

\begin{icmlauthorlist}
\icmlauthor{Koustuv Sinha}{fb,mcgill,mila}
\icmlauthor{Shagun Sodhani}{fb}
\icmlauthor{Joelle Pineau}{fb,mcgill,mila}
\icmlauthor{William L. Hamilton}{mcgill,mila}
\end{icmlauthorlist}

\icmlaffiliation{fb}{Facebook AI Research, Montreal, Canada}
\icmlaffiliation{mcgill}{School of Computer Science, McGill University, Montreal, Canada}
\icmlaffiliation{mila}{Montreal Institute of Learning Algorithms (Mila)}

\icmlcorrespondingauthor{Koustuv Sinha}{koustuv.sinha@mail.mcgill.ca}

\icmlkeywords{Machine Learning, ICML}

\vskip 0.3in
]




\section{GraphLog}
\label{sec:graph_log_app}

\subsection{Extended Terminology}
\label{sec:terminology_app}

In this section, we extend the terminology introduced in Section \ref{sec:terminology}. A set of relations is said to be \textit{Invertible} if



\begin{equation}\label{eq:invertible_rule}
    \forall r_i \in R, \exists r_j \in R \mid \{\forall u,v \in V_G : (u \rightarrow_{r_i} v)  \Rightarrow   (v \rightarrow_{r_j} u)\}.
\end{equation}

i.e. for all relations in $R$, there exists a relation in $R$ such that for all node pairs $(u, v)$ in the graph, if there exists an edge $u \rightarrow_{r_i} v$ then there exists another edge $v \rightarrow_{r_j} u$. Invertible relations are useful in determining the \textit{inverse} of a clause, where the directionality of the clause is flipped along with the ordering of the elements in the conjunctive clause. For example, the \textit{inverse} of Equation \ref{eq:rule} will be of the form:

\begin{equation}\label{eq:inv_rule}
    \exists z \in V_G : (v \rightarrow_{\hat{r_j}} z) \land  (z \rightarrow_{\hat{r_i}} u) \Rightarrow   (v \rightarrow_{\hat{r_k}} u) 
\end{equation}


\subsection{Dataset Generation}
\label{sec:dataset_generation_app}

This section follows up on the discussion in Section \ref{sec:dataset_generation}. We describe all the steps involved in the dataset generation process.

\xhdr{Rule Generation} In Algorithm \ref{alg:rule_generation}, we describe the complete process of generating rules in \name\ . We require the set of $K$ relations, which we use to sample the rule set $\mathcal{R}$. We mark some rules as being Invertible Rules (Section \ref{sec:terminology_app}). Then, we iterate through all possible combinations of relations in DataLog format to sample possible candidate rules. We impose two constraints on the candidate rule: \textbf{(i)} No two rules in $\mathcal{R}$ can have the same body. This ensures consistency between the rules. \textbf{(ii)} Candidate rules cannot have common relations among the \textit{head} and \textit{body}. 
This ensures absence of cycles. We also add the inverse rule of our sampled candidate rule and check the same consistencies again. We employ two types of unary Horn clauses to perform the closure of the available rules and to check the consistency of the different rules in $\mathcal{R}$.
Using this process, we ensure that all generated rules are sound and consistent with respect to $\mathcal{R}$.

\xhdr{World Sampling} 
\label{sec:world_sampling_app}
From the set of rules in $\mathcal{R}$, we partition rules into buckets for different worlds (Algorithm \ref{alg:rule_partition}). We use a simple policy of bucketing via a sliding window of width $w$ with stride $s$, to classify rules pertaining to each world.  For example, two such consecutive worlds can be generated as $\mathcal{R}^t = [\mathcal{R}_i \ldots \mathcal{R}_{i+w}]$ and $\mathcal{R}^{t+1} = [\mathcal{R}_{i+s} \ldots \mathcal{R}_{i+w+s}]$. (Algorithm \ref{alg:rule_partition}) We randomly permute $\mathcal{R}$ before bucketing in-order.

\xhdr{Graph Generation} This is a two-step process where first we sample a \textit{world graph} (Algorithm \ref{alg:world_graph}) and then we sample individual graphs from the \textit{world graph} (Algorithm \ref{alg:sample_graph}). Given a set of rules $\mathcal{R_S}$, in the first step, we recursively sample and apply rules in $\mathcal{R_S}$ to generate a \textit{relation graph} called \textit{world graph}. This sampling procedure enables us to create a diverse set of \textit{world graphs} by considering only certain subsets (of $\mathcal{R}$) during sampling. By controlling the extent of overlap between the subsets of $\mathcal{R}$ (in terms of the number of rules that are common across the subsets), we can precisely control the \textit{similarity} between the different \textit{world graphs}. 

In the second step (Algorithm \ref{alg:sample_graph}), the \textit{world graph} is used to sample a set of graphs $G_W^S = \{g_1, \cdots g_N\}$. A graph $g_i$ is sampled from $G_W$ by sampling a pair of nodes $(u, v)$ from $G_W$ and then by sampling a \textit{resolution path} $p_{G_W}^{u, v}$. The edge $u \rightarrow_{r_i} v$ provides the target relation that the learning model has to predict. Since the \textit{relation} for the edge 
$u \rightarrow_{r_i} v$ can be \textit{resolved} by composing the relations along the \textit{resolution} path, the relation prediction task tests for the compositional generalization abilities of the models. We first sample all possible resolution paths and get their individual descriptors $D_i$, which we split in training, validation and test splits. We then construct the training, validation and testing graphs by first adding all edges of an individual $D_i$ to the corresponding graph $g_i$, and then sampling neighbors of $p_{g_i}$. Concretely, we use Breadth First Search (BFS) to sample the neighboring subgraph of each node $u \in p_{g_i}$ with a decaying selection probability $\gamma$. This allows us to create diverse input graphs while having precise control over its resolution by its descriptor $D_i$. Splitting dataset over these descriptor paths ensures inductive generalization.


\begin{algorithm}[tb]
\caption{Rule Generator}
\label{alg:rule_generation}
\begin{algorithmic}
\STATE \textbf{Input:} Set of $K$ relations $\{r_i\}_K, K > 0$
\STATE Define an empty rule set $\mathcal{R}$
\STATE Populate Invertible Rules, $r_i \implies \hat{r_i}$, add to $\mathcal{R}$
\FORALL{$r_i \in \{r_i\}_K$}
\FORALL{$r_j \in \{r_i\}_K$}
\FORALL{$r_k \in \{r_i\}_K$}
\STATE Define candidate rule $t:[r_i, r_j] \implies r_k$
\IF{Cyclical rule, i.e. $r_i == r_k$ OR $r_j == r_k$}
\STATE Reject rule
\ENDIF
\IF{$t[body] \not\in \mathcal{R}$}
    \STATE Add $t$ to $\mathcal{R}$
    \STATE Define \textit{inverse} rule $t_{inv} : [\hat{r_j}, \hat{r_i}] \implies \hat{r_k}$
    \IF{$t_{inv}[body] \not\in \mathcal{R}$}
        \STATE Add $t_{inv}$ to $\mathcal{R}$
    \ELSE
        \STATE Remove rule having body $t_{inv}[body]$ from $\mathcal{R}$
    \ENDIF
\ENDIF
\ENDFOR
\ENDFOR
\ENDFOR
\STATE Check and remove any further cyclical rules.
\end{algorithmic}
\end{algorithm}
\begin{algorithm}
\caption{Partition rules into overlapping sets}
\label{alg:rule_partition}
\begin{algorithmic}
\REQUIRE Rule Set $\mathcal{R_S}$
\REQUIRE Number of worlds $n_w > 0$
\REQUIRE Number of rules per world $w > 0$
\REQUIRE Overlapping increment stride $s > 0$
\FOR{$i=0;i<|\mathcal{R_S}| - w;$}
    \STATE $\mathcal{R}_i = \mathcal{R_S}[i;i+w]$
    \STATE $i = i + s$
\ENDFOR
\end{algorithmic}
\end{algorithm}
\begin{algorithm}[tb]
\caption{World Graph Generator}
\label{alg:world_graph}
\begin{algorithmic}
\REQUIRE Set of relations $\{r_i\}_K, K > 0$
\REQUIRE Set of rules derived from $\{r_i\}_K$, $|\mathcal{R}| > 0$
\REQUIRE Set rule selection probability gamma $\gamma = 0.8$
\STATE Set rule selection probability $P[\mathcal{R}[i]] = 1, \forall i \in |\mathcal{R}|$
\REQUIRE Maximum number of expansions $s \ge 2 $
\REQUIRE Set of available nodes $N$, s.t. $|N| \ge 0$
\REQUIRE Number of cycles of generation $c \ge 0$
\STATE Set \textit{WorldGraph} set of edges $G_m = \emptyset$
\WHILE{$|N| > 0$ or $c > 0$}
\STATE Randomly choose an expansion number for this cycle: $\text{steps} = \texttt{rand}(2,s)$
\STATE Set added edges for this cycle $E_c = \emptyset$
\FORALL{step in steps}
\IF{step = 0}
\STATE With uniform probability, either:
    \STATE Sample $r_t$ from $\mathcal{R_S}[head]$ and sample $u,v \in N$ without replacement, OR 
    \STATE Sample an edge $(u, r_t, v)$ from $G_m$
\STATE Add $(u,r_t, v)$ to $E_c$ and $G_m$
\ELSE
    \STATE Sample an edge $(u, r_t, v)$ from $E_c$
\ENDIF
\STATE Sample a rule $\mathcal{R}[i]$ from $\mathcal{R}$ following $P$ s.t. $[r_i, r_j] \implies r_t$
\STATE $P[\mathcal{R}[i]] = P[\mathcal{R}[i]] * \gamma$
\STATE Sample a new node $y \in N$ without replacement
\STATE Add edge $(u,r_i,y)$ to $E_c$ and $G_m$
\STATE Add edge $(y,r_j,v)$ t $E_c$ and $G_m$
\ENDFOR
\IF{All rules in $\mathcal{R}$ is used atleast once}
\STATE Increment $c$ by 1
\STATE Reset rule selection probability $P[\mathcal{R}[i]] = 1, \forall i \in |\mathcal{R}|$
\ENDIF
\ENDWHILE
\end{algorithmic}
\end{algorithm}
\begin{algorithm}[tb]
\caption{Graph Sampler}
\label{alg:sample_graph}
\begin{algorithmic}
\REQUIRE Rule Set $\mathcal{R_S}$
\REQUIRE World Graph $G_m = (V_m, E_m)$
\REQUIRE Maximum Expansion length $e > 2$
\STATE Set Descriptor set $S = \emptyset$
\FORALL{$u,v \in E_m$}
\STATE Get all walks $Y_{(u,v)} \in G_m$ such that $|Y_{(u,v)}| \le e$
\STATE Get all descriptors $D_{Y_{(u,v)}}$ for all walks  $Y_{(u,v)}$
\STATE Add $D_{Y_{(u,v)}}$ to $S$
\ENDFOR
\STATE Set train graph set $G_{train} = \emptyset$
\STATE Set test graph set $G_{test} = \emptyset$
\STATE Split descriptors in train and test split, $S_{train}$ and $S_{test}$
\FORALL{$D_i \in S_{train}$ or $S_{test}$}
\STATE Set source node $u_s = D_i[0]$ and sink node $v_s = D_i[-1]$
\STATE Set prediction target $t = E_m[u_s][v_s]$
\STATE Set graph edges $g_i = \emptyset$
\STATE Add all edges from $D_i$ to $g_i$
\FORALL{$u,v \in D_i$}
\STATE Sample Breadth First Search connected nodes from $u$ and $v$ with decaying probability $\gamma$
\STATE Add the sampled edges to $g_i$
\ENDFOR
\STATE Remove edges in $g_i$ which create shorter paths between $u_s$ and $v_s$
\STATE Add $(g_i, u_s, v_s, t)$ to either $G_{train}$ or $G_{test}$
\ENDFOR
\end{algorithmic}
\end{algorithm}

\subsection{Computing Similarity}
\label{sec:dataset_similarity_app}

\name\ provides precise control for categorizing the similarity between different worlds by computing the overlap of the underlying rules. Concretely, the similarity between two worlds $W^i$ and $W^j$ is defined as $\text{Sim}(W^i, W^j) = |\mathcal{R}^i \cap \mathcal{R}^j|$, where $W_i$ and $W_j$ are the graph worlds and $\mathcal{R}^i$ and $\mathcal{R}^j$ are the set of rules associated with them. Thus \name\ enables various training scenarios - training on highly similar worlds or training on a mix of similar and dissimilar worlds. This fine grained control allows \name\ to mimic both in-distribution and out-of-distribution scenarios - during training and testing. It also enables us to precisely categorize the effect of multi-task pre-training when the model needs to adapt to novel worlds.

\subsection{Computing difficulty}
\label{sec:dataset_difficulty_app}
Recent research in multitask learning has shown evidence that models prioritize selection of difficult tasks over easy tasks while learning to boost the overall performance \cite{guo2018dynamic}. Thus, \name\ also provides a method to examine how pretraining on tasks of different difficulty level affects the adaptation performance. Due to the stochastic effect of partitioning of the rules, \name\ consists of datasets with varying range of difficulty. We use the supervised learning scores (Table \ref{tab:full_supervised}) as a proxy to determine the the relative difficulty of different datasets. We cluster the datasets such that tasks with prediction accuracy greater than or above 70\% are labeled as \textit{easy} difficulty, 50-70\% are labeled as \textit{medium} difficulty and below 50\% are labeled as \textit{hard} difficulty dataset. We find that the labels obtained by this criteria are consistent across the different models (Figure \ref{fig:inductive-reasoning}).



\section{Supervised learning on GraphLog}
\label{sec:full_world_stats_app}

We perform extensive experiments over \textit{all} the datasets available in \name\ (statistics given in  Table  \ref{tab:full_supervised}). We observe that in general, for the entire set of 57 worlds, the \texttt{GAT\_E-GAT} model performs the best. We observe that the relative difficulty (Section \ref{sec:dataset_difficulty_app}) of the tasks are highly correlated with the number of descriptors (Section \ref{sec:terminology_app}) available for each task. This shows that for a learner, a dataset with enough variety among the resolution paths of the graphs is relatively easier to learn compared to the datasets which has less variation.

\section{Multitask Learning}


\subsection{Multitask Learning on different data splits by difficulty}
\label{sec:mult_train_diff_app}

\begin{table}[h]
\centering
\resizebox{0.5\textwidth}{!}{%
\begin{tabular}{|ll|l|l|l|}
\hline
 &  & Easy & Medium & Difficult \\ \hline
$f_r$ & $f_c$ & Accuracy & Accuracy & Accuracy \\ \hline
GAT & E-GAT & \textbf{0.729} $\pm{0.05}$ & \textbf{0.586} $\pm{0.05}$ & 0.414 $\pm{0.07}$ \\
Param & E-GAT & 0.728 $\pm{0.05}$ & 0.574 $\pm{0.06}$ & 0.379 $\pm{0.06}$ \\
GCN & E-GAT & 0.713 $\pm{0.05}$ & 0.55 $\pm{0.06}$ & 0.396 $\pm{0.05}$ \\
GAT & RGCN & 0.695 $\pm{0.04}$ & 0.53 $\pm{0.03}$ & \textbf{0.421} $\pm{0.06}$ \\
Param & RGCN & 0.551 $\pm{0.08}$ & 0.457 $\pm{0.05}$ & 0.362 $\pm{0.05}$ \\
GCN & RGCN & 0.673 $\pm{0.05}$ & 0.514 $\pm{0.04}$ & 0.396 $\pm{0.06}$ \\ \hline
\end{tabular}%
}
\caption{Inductive performance on data splits marked by difficulty}
\label{tab:multitask_difficulty_inductive}
\end{table}

In Section \ref{sec:dataset_difficulty_app} we introduced the notion of \textit{difficulty} among the tasks available in \name\ . Here, we consider a set of experiments where we perform multitask training and inductive testing on the worlds bucketized by their relative difficulty (Table \ref{tab:multitask_difficulty_inductive}). We sample equal number of worlds from each difficulty bucket, and separately perform multitask training and testing. We evaluate the average prediction accuracy on the datasets within each bucket. We observe that the average multitask performance \textit{also} mimics the relative task difficulty distribution. We find \texttt{GAT-E-GAT} model outperforms other baselines in \textit{Easy} and \textit{Medium} setup, but is outperformed by \texttt{GAT-RGCN} model in the \textit{Difficult} setup. For each model, we used the same architecture and hyperparameter settings across the buckets. Optimizing individually for each bucket may improve the relative performance.

\subsection{Multitask Pre-training by task similarity}
\label{sec:mult_pre_train_sim_app}

\begin{figure}[h]
    \centering
    \includegraphics[width=0.5\textwidth]{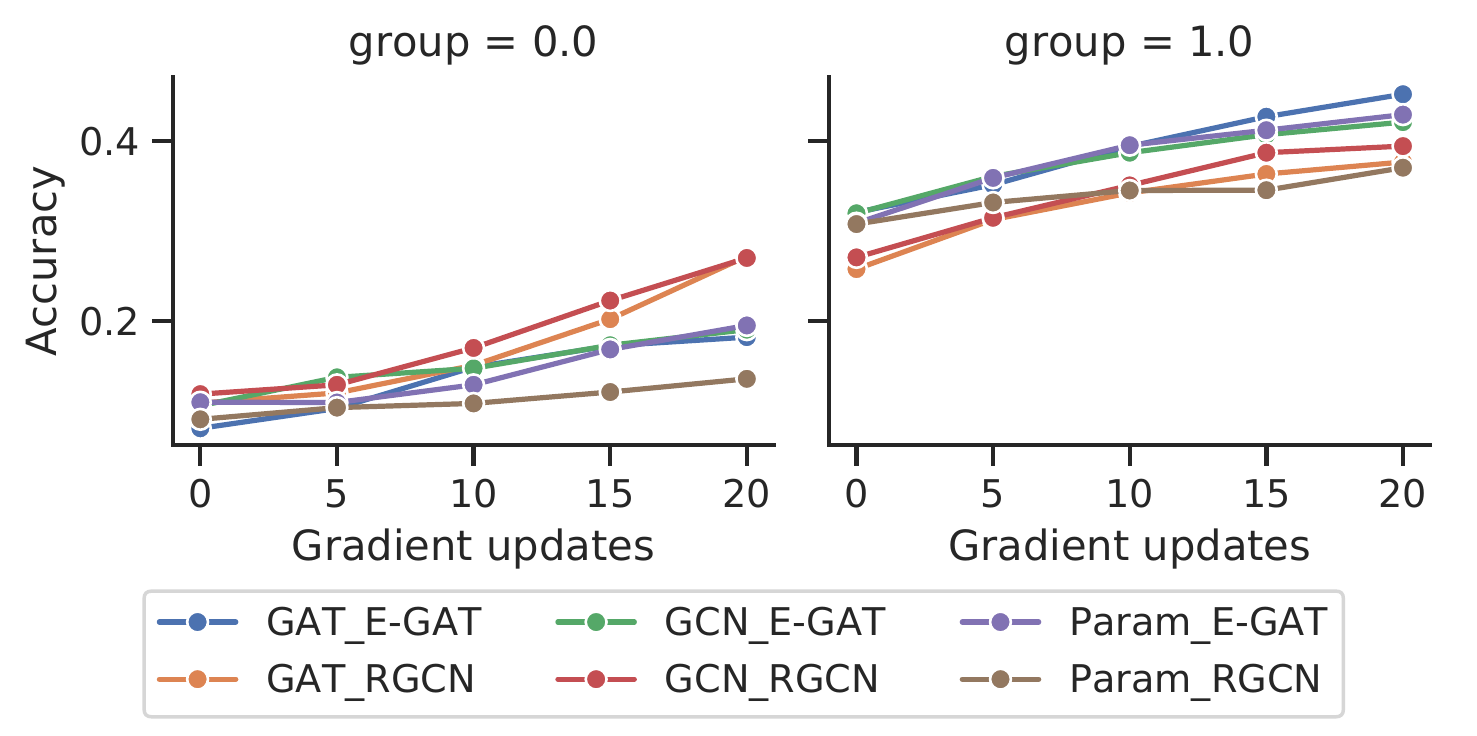}
    \caption{
    We perform fine-grained analysis of \textit{few shot} adaptation capabilities in Multitask setting. Group 0.0 and 1.0 corresponds to 0\% and 100\% similarity respectively. }
    \label{fig:multitask_fine_grained}
\end{figure}

In the main paper (Section \ref{sec:multi_task_training_setup}) we introduce the setup of performing multitask pre-training on \name\ datasets and adaptation on the datasets based on relative similarity. Here, we perform fine-grained analysis of \textit{few-shot} adapatation capabilities of the models. We analyze the adaptation performance in two settings - when the adaptation dataset has complete overlap of rules with the training datasets (\textit{group}=1.0) and when the adaptation dataset has zero overlap with the training datasets (\textit{group}=0.0). We find RGCN family of models with a graph based representation function has faster adaptation on the dissimilar dataset, with \texttt{GCN-RGCN} showing the fastest improvement. However on the similar dataset the models follow the ranking of the supervised learning experiments, with \texttt{GAT-EGAT} model adapting comparitively better.

\subsection{Multitask Pre-training by task difficulty}
\label{sec:mult_pre_train_diff_app}

\begin{table}[]
\centering
\resizebox{0.5\textwidth}{!}{%
\begin{tabular}{|ll|l|l|l|}
\hline
 &  & Easy & Medium & Difficult \\ \hline
$f_r$ & $f_c$ & Accuracy & Accuracy & Accuracy \\ \hline
GAT & E-GAT & 0.531 $\pm{0.03}$ & \textbf{0.569} $\pm{0.01}$ & \textbf{0.555} $\pm{0.04}$ \\
Param & E-GAT & 0.520 $\pm{0.02}$ & 0.548 $\pm{0.01}$ & 0.540 $\pm{0.01}$ \\
GCN & E-GAT & \textbf{0.555} $\pm{0.01}$ & 0.561 $\pm{0.02}$ & 0.558 $\pm{0.01}$ \\
GAT & RGCN & 0.502 $\pm{0.02}$ & 0.532 $\pm{0.01}$ & 0.532 $\pm{0.01}$ \\
Param & RGCN & 0.535 $\pm{0.01}$ & 0.506 $\pm{0.04}$ & 0.539 $\pm{0.04}$ \\
GCN & RGCN & 0.481 $\pm{0.02}$ & 0.516 $\pm{0.02}$ & 0.520 $\pm{0.01}$ \\ \hline
\multicolumn{2}{|c|}{Mean} & \multicolumn{1}{c|}{0.521} & \multicolumn{1}{c|}{\textbf{0.540}} & \multicolumn{1}{c|}{0.539} \\ \hline
\end{tabular}%
}
\caption{Convergence performance on 3 held out datasets when pre-trained on easy, medium and hard training datasets}
\label{tab:multitask_difficulty_convergence}
\end{table}

\begin{figure}
    \centering
    \includegraphics[width=0.5\textwidth]{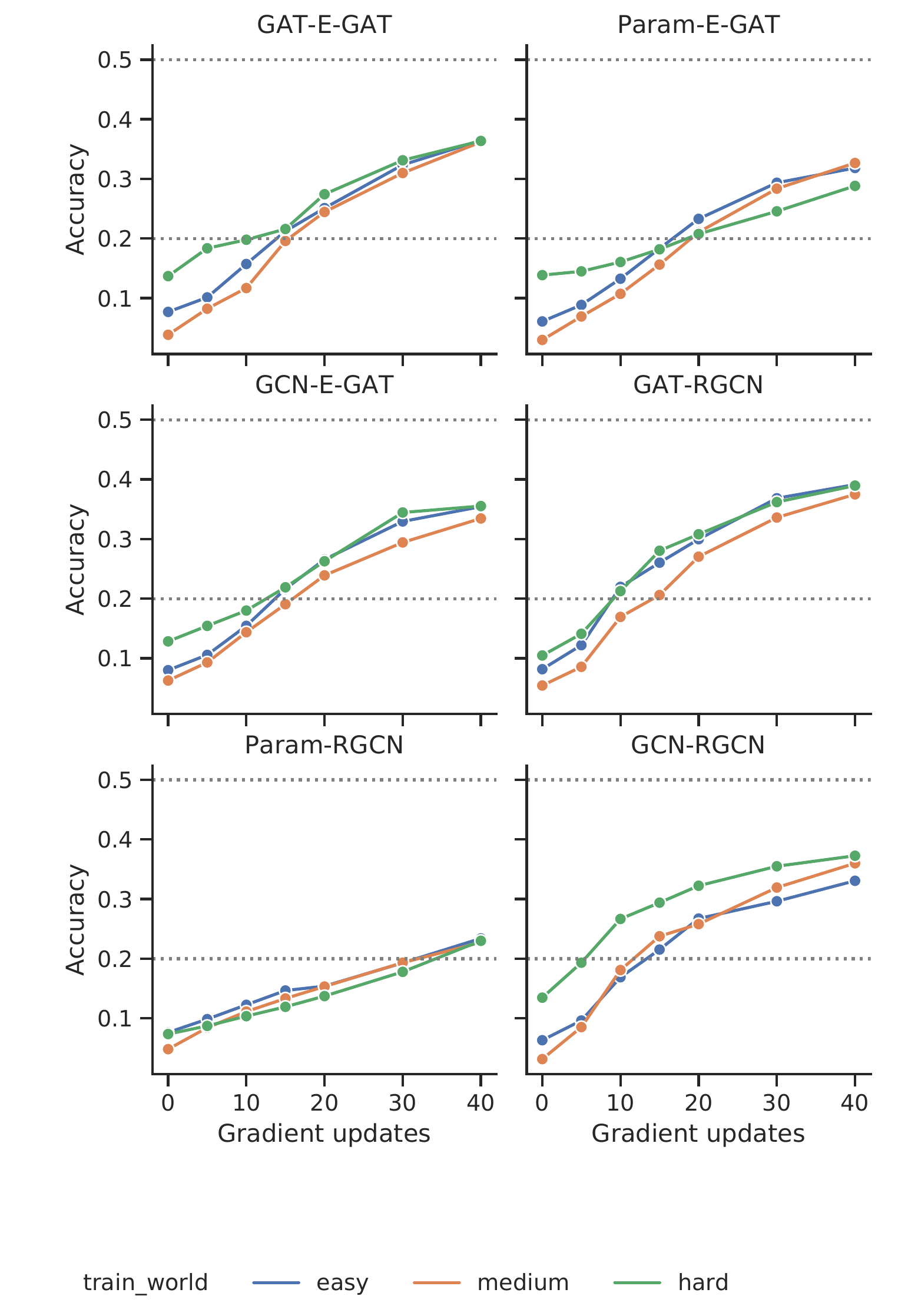}
    \caption{
    We evaluate the effect of $k$-shot adaptation on held out datasets when pre-trained on \textit{easy}, \textit{medium} and \textit{hard} training datasets, among the different model architectures. Here, $k$ ranges from 0 to 40.}
    \label{fig:effect_of_difficulty_mult}
\end{figure}

Using the notion of \textit{difficulty} introduced in Section \ref{sec:dataset_difficulty_app}, we perform the suite of experiments to evaluate the effect of pre-training on \textit{Easy}, \textit{Medium} and \textit{Difficult} datasets. Interestingly, we find the performance on convergence is better on Medium and Hard datasets on pre-training, compared to the Easy dataset (Table \ref{tab:multitask_difficulty_convergence}). This behaviour is also mirrored in k-shot adaptation performance (Figure \ref{fig:effect_of_difficulty_mult}), where pre-training on Hard dataset provides faster adaptation performance on 4/6 models.

\section{Continual Learning}

\begin{figure}
    \centering
    \includegraphics[width=0.5\textwidth]{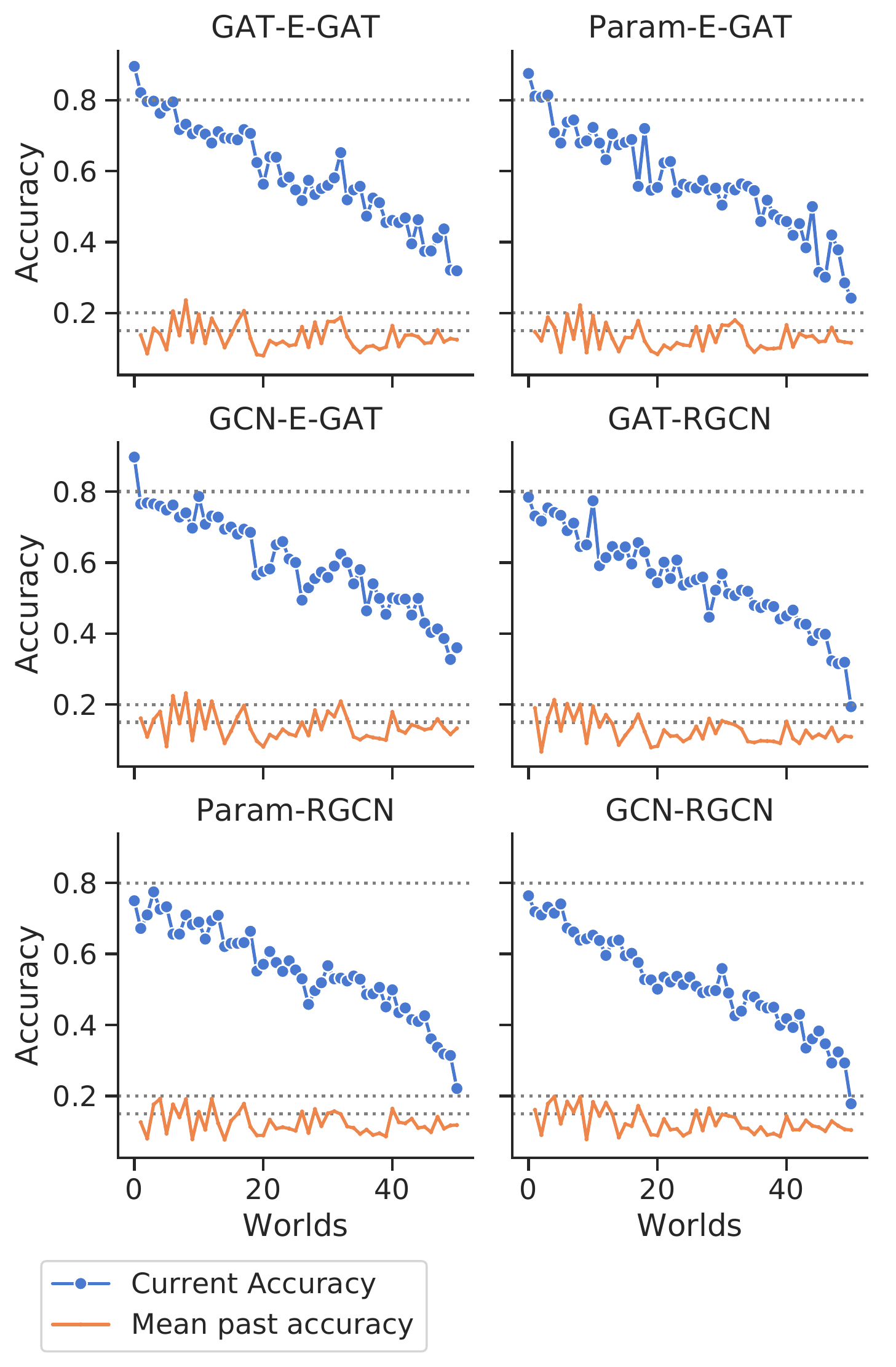}
    \caption{
    Curriculum Learning strategy in Continual Learning setup of GraphLog.}
    \label{fig:effect_of_curriculum}
\end{figure}

\begin{figure}
    \centering
    \includegraphics[width=0.5\textwidth]{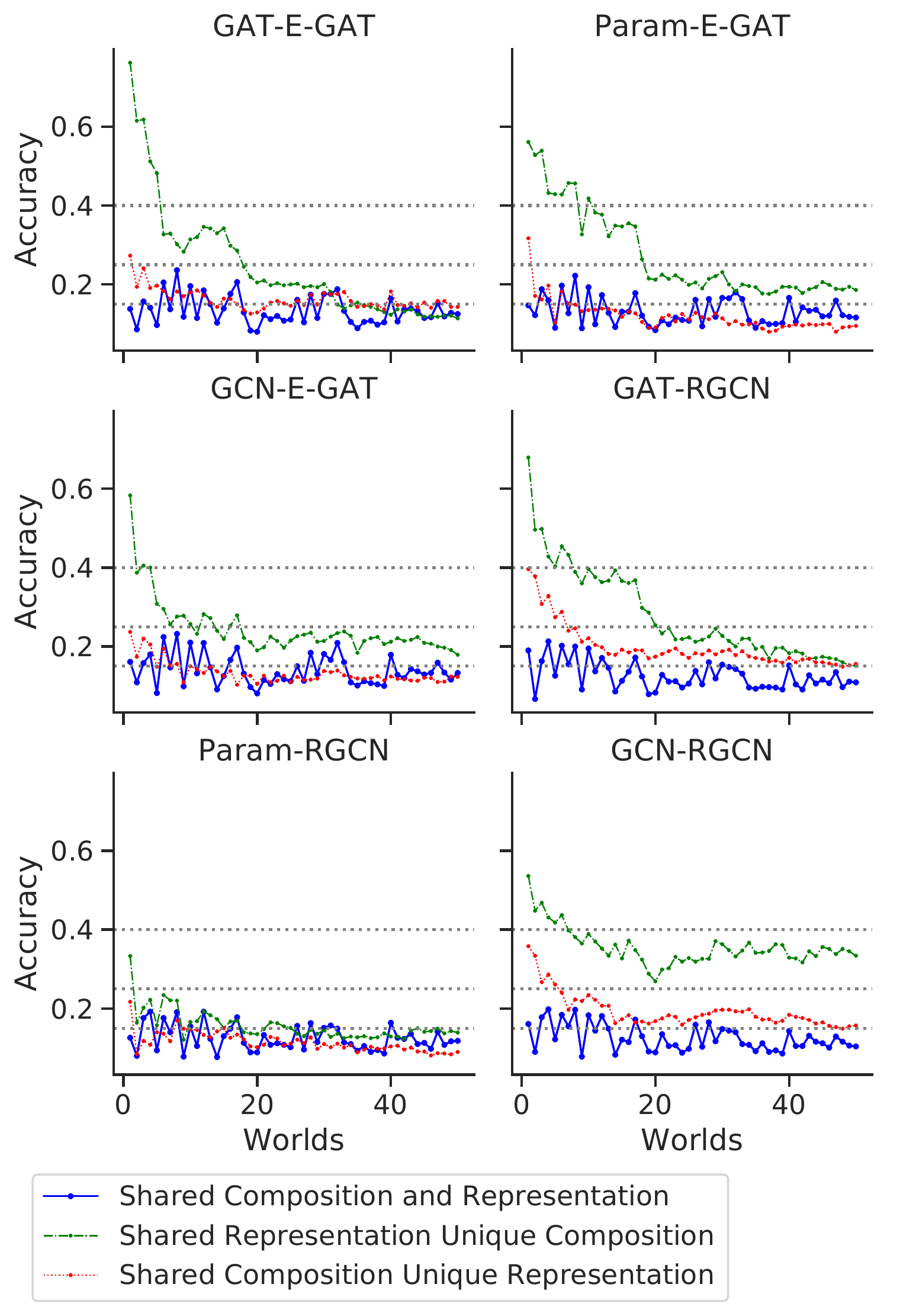}
    \caption{
    Curriculum Learning strategy in Continual Learning setup of GraphLog, when either the composition function or the representation function is shared for all worlds.}
    \label{fig:effect_of_curriculum_comp_rep}
\end{figure}

A natural question arises following our continual learning experiments in Section \ref{sec:continual_learning} : does the \textit{order} of difficulty of the worlds matter? Thus, we perform an experiment following Curriculum Learning \cite{bengio2009curriculum} setup, where the order of the worlds being trained is determined by their relative difficulty (which is determined by the performance of models in supervised learning setup, Table \ref{tab:full_supervised}, i.e., we order the worlds from easier worlds to harder worlds). We observe that while the current task accuracy follows the trend of the difficulty of the worlds (Figure \ref{fig:effect_of_curriculum}), the mean of past accuracy is significantly worse. This suggests that a curriculum learning strategy might not be optimal to learn graph representations in a continual learning setting. We also performed the same experiment with sharing only the composition and representation functions (Figure \ref{fig:effect_of_curriculum_comp_rep}), and observe similar trends where sharing the representation function reduces the effect of catastrophic forgetting.

\begin{table*}[]
\centering
\resizebox{\textwidth}{!}{%
\begin{tabular}{lrrlrrrlrrrrrr}
\toprule
World ID &  NC &    ND &  Split &  ARL &  AN &  AE & D &  M1 &  M2 &  M3 &  M4 &  M5 &  M6 \\
\midrule
  rule\_0 &         17 &   286 &  train &                       4.49 &         15.487 &         19.295 &       Hard &      0.481 &      0.500 &        0.494 &     0.486 &     0.462 &       0.462 \\
  rule\_1 &         15 &   239 &  train &                       4.10 &         11.565 &         13.615 &       Hard &      0.432 &      0.411 &        0.428 &     0.406 &     0.400 &       0.408 \\
  rule\_2 &         17 &   157 &  train &                       3.21 &          9.809 &         11.165 &       Hard &      0.412 &      0.357 &        0.373 &     0.347 &     0.347 &       0.319 \\
  rule\_3 &         16 &   189 &  train &                       3.63 &         11.137 &         13.273 &       Hard &      0.429 &      0.404 &        0.473 &     0.373 &     0.401 &       0.451 \\
  rule\_4 &         16 &   189 &  train &                       3.94 &         12.622 &         15.501 &     Medium &      0.624 &      0.606 &        0.619 &     0.475 &     0.481 &       0.595 \\
  rule\_5 &         14 &   275 &  train &                       4.41 &         14.545 &         18.872 &       Hard &      0.526 &      0.539 &        0.548 &     0.429 &     0.461 &       0.455 \\
  rule\_6 &         16 &   249 &  train &                       5.06 &         16.257 &         20.164 &       Hard &      0.528 &      0.514 &        0.536 &     0.498 &     0.495 &       0.476 \\
  rule\_7 &         17 &   288 &  train &                       4.47 &         13.161 &         16.333 &     Medium &      0.613 &      0.558 &        0.598 &     0.487 &     0.486 &       0.537 \\
  rule\_8 &         15 &   404 &  train &                       5.43 &         15.997 &         19.134 &     Medium &      0.627 &      0.643 &        0.629 &     0.523 &     0.563 &       0.569 \\
  rule\_9 &         19 &  1011 &  train &                       7.22 &         24.151 &         32.668 &       Easy &      0.758 &      0.744 &        0.739 &     0.683 &     0.651 &       0.623 \\
 rule\_10 &         18 &   524 &  train &                       5.87 &         18.011 &         22.202 &     Medium &      0.656 &      0.654 &        0.663 &     0.596 &     0.563 &       0.605 \\
 rule\_11 &         17 &   194 &  train &                       4.29 &         11.459 &         13.037 &     Medium &      0.552 &      0.525 &        0.533 &     0.445 &     0.456 &       0.419 \\
 rule\_12 &         15 &   306 &  train &                       4.14 &         11.238 &         12.919 &       Easy &      0.771 &      0.726 &        0.603 &     0.511 &     0.561 &       0.523 \\
 rule\_13 &         16 &   149 &  train &                       3.58 &         11.238 &         13.549 &       Hard &      0.453 &      0.402 &        0.419 &     0.347 &     0.298 &       0.344 \\
 rule\_14 &         16 &   224 &  train &                       4.14 &         11.371 &         13.403 &       Hard &      0.448 &      0.457 &        0.401 &     0.314 &     0.318 &       0.332 \\
 rule\_15 &         14 &   224 &  train &                       3.82 &         12.661 &         15.105 &       Hard &      0.494 &      0.423 &        0.501 &     0.402 &     0.397 &       0.435 \\
 rule\_16 &         16 &   205 &  train &                       3.59 &         11.345 &         13.293 &       Hard &      0.318 &      0.332 &        0.292 &     0.328 &     0.306 &       0.291 \\
 rule\_17 &         17 &   147 &  train &                       3.16 &          8.163 &          8.894 &       Hard &      0.347 &      0.308 &        0.274 &     0.164 &     0.161 &       0.181 \\
 rule\_18 &         18 &   923 &  train &                       6.63 &         25.035 &         33.080 &       Easy &      0.700 &      0.680 &        0.713 &     0.650 &     0.641 &       0.618 \\
 rule\_19 &         16 &   416 &  train &                       6.10 &         17.180 &         20.818 &       Easy &      0.790 &      0.774 &        0.777 &     0.731 &     0.729 &       0.702 \\
 rule\_20 &         20 &  2024 &  train &                       8.63 &         34.059 &         45.985 &       Easy &      0.830 &      0.799 &        0.854 &     0.756 &     0.741 &       0.750 \\
 rule\_21 &         13 &   272 &  train &                       4.58 &         10.559 &         11.754 &     Medium &      0.621 &      0.610 &        0.632 &     0.531 &     0.516 &       0.580 \\
 rule\_22 &         17 &   422 &  train &                       5.21 &         16.540 &         20.681 &     Medium &      0.586 &      0.593 &        0.628 &     0.530 &     0.506 &       0.573 \\
 rule\_23 &         15 &   383 &  train &                       4.97 &         17.067 &         21.111 &       Hard &      0.508 &      0.522 &        0.493 &     0.455 &     0.473 &       0.476 \\
 rule\_24 &         18 &   879 &  train &                       6.33 &         21.402 &         26.152 &       Easy &      0.706 &      0.704 &        0.743 &     0.656 &     0.641 &       0.638 \\
 rule\_25 &         15 &   278 &  train &                       3.84 &         11.093 &         12.775 &       Hard &      0.424 &      0.419 &        0.382 &     0.358 &     0.345 &       0.412 \\
 rule\_26 &         15 &   352 &  train &                       4.71 &         14.157 &         17.115 &     Medium &      0.565 &      0.534 &        0.532 &     0.466 &     0.461 &       0.499 \\
 rule\_27 &         16 &   393 &  train &                       4.98 &         14.296 &         16.499 &       Easy &      0.713 &      0.714 &        0.722 &     0.632 &     0.604 &       0.647 \\
 rule\_28 &         16 &   391 &  train &                       4.82 &         17.551 &         21.897 &     Medium &      0.575 &      0.564 &        0.571 &     0.503 &     0.499 &       0.552 \\
 rule\_29 &         16 &   144 &  train &                       3.87 &         10.193 &         11.774 &       Hard &      0.468 &      0.445 &        0.475 &     0.325 &     0.336 &       0.389 \\
 rule\_30 &         17 &   177 &  train &                       3.51 &         10.270 &         11.764 &       Hard &      0.381 &      0.426 &        0.382 &     0.357 &     0.316 &       0.336 \\
 rule\_31 &         19 &   916 &  train &                       5.90 &         20.147 &         26.562 &       Easy &      0.788 &      0.789 &        0.770 &     0.669 &     0.674 &       0.641 \\
 rule\_32 &         16 &   287 &  train &                       4.66 &         16.270 &         20.929 &     Medium &      0.674 &      0.671 &        0.700 &     0.621 &     0.594 &       0.615 \\
 rule\_33 &         18 &   312 &  train &                       4.50 &         14.738 &         18.266 &     Medium &      0.695 &      0.660 &        0.709 &     0.710 &     0.679 &       0.668 \\
 rule\_34 &         18 &   504 &  train &                       5.00 &         15.345 &         18.614 &       Easy &      0.908 &      0.888 &        0.906 &     0.768 &     0.762 &       0.811 \\
 rule\_35 &         19 &   979 &  train &                       6.23 &         21.867 &         28.266 &       Easy &      0.831 &      0.750 &        0.782 &     0.680 &     0.700 &       0.662 \\
 rule\_36 &         19 &   252 &  train &                       4.66 &         13.900 &         16.613 &       Easy &      0.742 &      0.698 &        0.698 &     0.659 &     0.627 &       0.651 \\
 rule\_37 &         17 &   260 &  train &                       4.00 &         11.956 &         14.010 &       Easy &      0.843 &      0.826 &        0.826 &     0.673 &     0.698 &       0.716 \\
 rule\_38 &         17 &   568 &  train &                       5.21 &         15.305 &         20.075 &       Easy &      0.748 &      0.762 &        0.733 &     0.644 &     0.630 &       0.719 \\
 rule\_39 &         15 &   182 &  train &                       3.98 &         12.552 &         14.800 &       Easy &      0.737 &      0.642 &        0.635 &     0.592 &     0.603 &       0.587 \\
 rule\_40 &         17 &   181 &  train &                       3.69 &         11.556 &         14.437 &     Medium &      0.552 &      0.584 &        0.575 &     0.525 &     0.472 &       0.479 \\
 rule\_41 &         15 &   113 &  train &                       3.58 &         10.162 &         11.553 &     Medium &      0.619 &      0.601 &        0.626 &     0.490 &     0.468 &       0.470 \\
 rule\_42 &         14 &    95 &  train &                       2.96 &          8.939 &          9.751 &       Hard &      0.511 &      0.472 &        0.483 &     0.386 &     0.393 &       0.395 \\
 rule\_43 &         16 &   162 &  train &                       3.36 &         11.077 &         13.337 &     Medium &      0.622 &      0.567 &        0.579 &     0.473 &     0.482 &       0.437 \\
 rule\_44 &         18 &   705 &  train &                       4.75 &         15.310 &         18.172 &       Hard &      0.538 &      0.561 &        0.603 &     0.498 &     0.519 &       0.450 \\
 rule\_45 &         15 &   151 &  train &                       3.39 &          9.127 &         10.001 &     Medium &      0.569 &      0.580 &        0.592 &     0.535 &     0.524 &       0.524 \\
 rule\_46 &         19 &  2704 &  train &                       7.94 &         31.458 &         43.489 &       Easy &      0.850 &      0.820 &        0.828 &     0.773 &     0.762 &       0.749 \\
 rule\_47 &         18 &   647 &  train &                       6.66 &         22.139 &         27.789 &       Easy &      0.723 &      0.667 &        0.708 &     0.620 &     0.649 &       0.611 \\
 rule\_48 &         16 &   978 &  train &                       6.15 &         17.802 &         21.674 &       Easy &      0.812 &      0.798 &        0.812 &     0.772 &     0.763 &       0.753 \\
 rule\_49 &         14 &   169 &  train &                       3.41 &          9.983 &         11.177 &       Easy &      0.714 &      0.734 &        0.700 &     0.511 &     0.491 &       0.615 \\
 rule\_50 &         16 &   286 &  train &                       3.99 &         12.274 &         16.117 &     Medium &      0.651 &      0.653 &        0.656 &     0.555 &     0.583 &       0.570 \\
 rule\_51 &         16 &   332 &  valid &                       4.44 &         16.384 &         21.817 &       Easy &      0.746 &      0.742 &        0.738 &     0.667 &     0.657 &       0.689 \\
 rule\_52 &         17 &   351 &  valid &                       4.81 &         16.231 &         20.613 &     Medium &      0.697 &      0.716 &        0.754 &     0.653 &     0.655 &       0.670 \\
 rule\_53 &         15 &   165 &  valid &                       3.65 &         10.838 &         12.378 &       Hard &      0.458 &      0.464 &        0.525 &     0.334 &     0.364 &       0.373 \\
 rule\_54 &         13 &   303 &   test &                       5.25 &         13.503 &         15.567 &     Medium &      0.638 &      0.623 &        0.603 &     0.587 &     0.586 &       0.555 \\
 rule\_55 &         16 &   293 &   test &                       4.83 &         16.444 &         20.944 &     Medium &      0.625 &      0.582 &        0.578 &     0.561 &     0.528 &       0.571 \\
 rule\_56 &         15 &   241 &   test &                       4.40 &         14.010 &         16.702 &     Medium &      0.653 &      0.681 &        0.692 &     0.522 &     0.513 &       0.550 \\ \midrule
AGG  & 16.33 & 428.94 &  & 4.70 & 14.89 & 18.37 &   &  \textbf{0.618} / \textbf{26} &     0.603 / 10 &  0.611 / 20 & 0.530 / 1 & 0.526 / 0 & 0.539 / 0 \\
\bottomrule
\end{tabular}%
}
\caption{Results on Single-task supervised setup for all datasets in GraphLog. Abbreviations: \textbf{NC}: Number of Classes, \textbf{ND}: Number of Descriptors, \textbf{ARL}: Average Resolution Length, \textbf{AN}: Average number of nodes, \textbf{AE}: Average number of edges}, \textbf{D}: Difficulty, \textbf{AGG}: Aggregate Statistics. List of models considered : \textbf{M1}: GAT-EGAT, \textbf{M2}: GCN-E-GAT, \textbf{M3}: Param-E-GAT, \textbf{M4}: GAT-RGCN, \textbf{M5}: GCN-RGCN and \textbf{M6}: Param-RGCN. Difficulty is calculated by taking the scores of the model (M1) and partitioning the worlds according to their accuracy ($\ge 0.7$ = Easy, $\ge 0.54$ and $< 0.7$ = Medium, and $< 0.54$ =  Hard). We provide both the mean of the raw accuracy scores for all models, as well as the number of times the model is ranked first in all the tasks.
\label{tab:full_supervised}
\end{table*}

\section{Hyperparameters and Experimental Setup}
\label{sec:hyperparam_app}

In this section, we provide detailed hyperparameter settings for both models and dataset generation for the purposes of reproducibility. The codebase and dataset used in the experiments are attached with the Supplementary materials, and will be made public on acceptance.

\subsection{Dataset Hyperparams}
\label{sec:dataset_hyp_app}

We generate \name\ with 20 relations or classes ($K$), which results in 76 rules in $\mathcal{R_S}$ after consistency checks. For unary rules, we specify half of the relations to be symmetric and other half to have their invertible relations. To split the rules for individual worlds, we choose the number of rules for each world $w=20$ and stride $s=1$, and end up with 57 worlds $\mathcal{R}_0 \ldots \mathcal{R}_{56}$. For each world $\mathcal{R}_i$, we generate 5000 training, 1000 testing and 1000 validation graphs. 

\subsection{Model Hyperparams}
\label{sec:model_hyp_app}

For all models, we perform hyper-parameter sweep (grid search) to find the optimal values based on the validation accuracy. For all models, we use the relation embedding and node embedding to be 200 dimensions. We train all models with Adam optimizer with learning rate 0.001 and weight decay of 0.0001. For supervised setting, we train all models for 500 epochs, and we add a scheduler for learning rate to decay it by 0.8 whenever the validation loss is stagnant for 10 epochs. In multitask setting, we sample a new task every epoch from the list of available tasks. Here, we run all models for 2000 epochs when we have the number of tasks $\le 10$. For larger number of tasks (Figure \ref{fig:mult_task_large}), we train by proportionally increasing the number of epochs compared to the number of tasks. (2k epochs for 10 tasks, 4k epochs for 20 tasks, 6k epochs for 30 tasks, 8k epochs for 40 tasks and 10k epochs for 50 tasks). For continual learning experiment, we train each task for 100 epochs for all models. No learning rate scheduling is used for either multitask or continual learning experiments. Individual model hyper-parameters are as follows:

\begin{itemize}
    \item Representation functions :
    \begin{itemize}
        \item \texttt{GAT} : Number of layers = 2, Number of attention heads = 2, Dropout = 0.4
        \item \texttt{GCN} : Number of layers = 2, with symmetric normalization and bias, no dropout
    \end{itemize}
    \item Composition functions:
    \begin{itemize}
        \item \texttt{E-GAT}: Number of layers = 6, Number of attention heads = 2, Dropout = 0.4
        \item \texttt{RGCN}: Number of layers = 2, no dropout, with bias.
    \end{itemize}
\end{itemize}


\end{document}